\pdfoutput=1
\RequirePackage[2020-02-02]{latexrelease}
\documentclass[11pt]{article}

\usepackage[]{emnlp2021}
\usepackage{times}
\usepackage{latexsym}

\usepackage[T1]{fontenc}
\usepackage[utf8]{inputenc}
\usepackage{xcolor}

\usepackage{booktabs}
\usepackage{times}
\usepackage{latexsym}
\usepackage{natbib}
\usepackage{amsmath}
\usepackage{bbm}
\usepackage{graphicx}

\usepackage{color}
\usepackage{amsmath}
\usepackage{amssymb}
\usepackage{booktabs}
\usepackage{graphicx}
\usepackage{hyperref}
\usepackage{relsize}
\usepackage{latexsym}
\usepackage{paralist}
\usepackage{times}
\usepackage[printwatermark]{xwatermark}

\usepackage{url}
\usepackage{xspace}
\usepackage{diagbox}
\usepackage{multirow}
\usepackage{makecell}
\usepackage{bigdelim}
\usepackage{mathtools}

\usepackage{booktabs}
\usepackage{microtype}

\usepackage{xspace}
\newcommand{\cc}{Common Crawl\xspace}
\newcommand{\wikipedia}{Wikipedia\xspace}

\newcommand{\openwebtext}{OpenWebText\xspace}
\newcommand{\books}{books\xspace}
\newcommand{\openaifilter}{\textsc{GPT-3} quality filter\xspace}
\newcommand{\openaifiltercaps}{\textsc{GPT-3} Quality Filter\xspace}

\newcommand{\booksthree}{Books3\xspace}

\newcommand{\gptthree}{\textsc{GPT-3}\xspace}

\newcommand{\schoolnewspapers}{\textsc{U.S.~School News}\xspace}

\newcommand{\phighquality}{$P(\text{high quality})$}

\usepackage{dcolumn}
\newcolumntype{d}[1]{D{.}{.}{#1}}  % define "d" column type

\title{Whose Language Counts as High Quality? \\ Measuring Language Ideologies in Text Data Selection}

\author{
    Suchin Gururangan$^{\dagger}$ \quad
	Dallas Card$^{\diamondsuit}$ \quad 
	\bf Sarah K. Dreier$^{\heartsuit}$ \quad 
	\bf Emily K. Gade$^{\clubsuit}$ \quad  \\
	\bf Leroy Z. Wang$^{\dagger}$ \quad 
	\bf Zeyu Wang$^{\dagger}$ \quad 
	\bf Luke Zettlemoyer$^{\dagger}$ \quad
	\bf Noah A. Smith$^{\dagger\spadesuit}$ \quad \\
	$^\dagger$University of Washington   \quad  $^\diamondsuit$ University of Michigan \quad $^\heartsuit$University of New Mexico \\
	$^\clubsuit$Emory University \quad
	$^\spadesuit$Allen Institute for AI \\
    \tt \{sg01, zwan4, lsz, nasmith\}@cs.washington.edu dalc@umich.edu \\ 
    \tt skdreier@unm.edu emily.gade@emory.edu lryw@uw.edu
}

\begin{document}
\maketitle
%\begin{abstract}
%\end{abstract}

\begin{abstract}
% \nascomment{propose to avoid ``pretraining'' since that centers the downstream task model (and there are none of those in this paper).  from a language model's perspective, it's just data or training data!  I suggest the title refer to ``text data selection'' not ``pretraining data selection''}
    % LATEST ABSTRACT
    % \dallas{Attempt at a rewrite; may need to iterate more still}
    % Language models increasingly rely on web content for diverse pretraining data. However, the use of web content requires some level of filtering, with particular datasets often serving as the anchor for identifying ``high quality'' text. As such, despite their extensive biases, resources such as Wikipedia, books, and newswire are positioned as the gold standard of quality. Using a new dataset of scraped U.S. high school newspaper articles---written by students from across the country---we investigate whose language is preferred by the quality filter used for GPT-3. We find that newspapers from larger schools, located in wealthier, educated, and urban ZIP codes are more likely to be classified as high quality. We further demonstrate that the filter's measurement of quality is unaligned with other sensible metrics, such as factuality or literary acclaim. Overall, we argue that privileging any corpus as high quality entails an implicit language ideology and more care is needed in constructing training corpora for language models, with better transparency and justification for the inclusion or exclusion of certain text.

    Language models increasingly rely on massive web dumps for diverse text data. However, these sources are rife with undesirable content. As such, resources like Wikipedia, books, and newswire often serve as anchors for automatically selecting web text most suitable for language modeling, a process typically referred to as \emph{quality filtering}. Using a new dataset of U.S. high school newspaper articles---written by students from across the country---we investigate whose language is preferred by the quality filter used for \gptthree. We find that newspapers from larger schools, located in wealthier, educated, and urban ZIP codes are more likely to be classified as high quality. We then demonstrate that the filter's measurement of quality is unaligned with other sensible metrics, such as factuality or literary acclaim. We argue that privileging any corpus as high quality entails a language ideology, and more care is needed to construct training corpora for language models, with better transparency and justification for the inclusion or exclusion of various texts.

\end{abstract}

% \suchin{Does the abstract/title need to be a bit broader, based on the pretraining data survey? Perhaps this is not just about quality filters, but that our pretraining data is generally not representative. }

% \suchin{Noah: curate, don't filter}

% \suchin{Noah: don't aim for better representation, cuz impossible}
% \suchin{Noah: we should argue that we should move to a "curation" mode when collecting pretraining data, rather than our current "laissez faire mode" of sample as much data and then filter. As we show in section 2, many of these datasets are naturally zipfian, so filtering will maintain/amplify that distribution}
% \suchin{Noah: we may argue that there is no such thing as a fully representative corpus (he agrees with this position), but that will be controversial and so we should be careful with that}

% \suchin{Noah: likes the idea of better employing (socio)linguists to help curate corpora at scale (e.g., we need better automated methods to detect the prevalence of dialects or other community-specific language)}

% \suchin{I really like using the term "language ideology" as what quality filters are imposing, since it's grounded in sociolinguistic literature. Maybe it's something we should elevate more in the intro. If we do this, we should explicitly introduce/describe it in the intro/section 2.}

%\suchin{From Noah/Luke: Change "positionality" in title (and elsewhere in draft) to something else.}
\section{Introduction}

The language models central to modern NLP  are trained on large Internet corpora, typically gathered from community resources (e.g., \wikipedia; \citealt{liu2019roberta}) or web dumps (e.g., WebText, \cc; \citealt{radford2019language}, \citealt{brown2020language}).  The selection of texts impacts every research or deployed NLP system that builds on these models. Yet there is rarely any explicit justification for why various texts were included.

Web dumps like \cc offer the promise of more diverse text than what is available in curated resources. However, much of the web consists of frequently replicated boilerplate (e.g., privacy policies), code (e.g., HTML and Javascript), pornography, hate speech, and more. Automated approaches, typically referred to as \textbf{quality filters},\footnote{
We note that the term \emph{quality} is often ill-defined in the NLP literature. For example, \citet{brown2020language} and \citet{Wenzek2020CCNetEH} refer to ``high-quality text'' or ``high-quality sources''---both citing Wikipedia as an example---but without explaining precisely what is meant.
%These are both clearly good-faith efforts driven primarily by empirical concerns, but they illustrate the often unstated assumption that quality is .
%they illustrate a common assumption that certain resource(s) represent ideal or reference text, in terms of content and style, for language modeling.
%In this work, we demonstrate that such value judgments have implicit and as-yet undocumented effects on whose language is more likely to be included in the corpus.
} are  often applied in an effort to remove this undesirable content from training data. These filters include code removers  \citep{gao2020pile}, heuristics \citep{rae2021scaling}, stopwords \citep{raffel2020exploring}, and classifiers \citep{brown2020language, Wenzek2020CCNetEH}.

Although quality filtering is often treated as a relatively neutral preprocessing step, it necessarily implies a value judgment: which data is assumed to be of sufficiently high quality to be included in the training corpus?  More concretely, when a quality filter is a classifier trained on instances assumed to be of high (and low) quality, the selection of those examples will impact the language model and any downstream technology that uses it.  Many filters use \wikipedia, \books, and newswire to represent high quality text.  But what texts are excluded as a result?  Because natural language varies with social and demographic variables \citep[][\emph{inter alia}]{10.2307/455300, eckert1989jocks,labov_2006,blodgett-etal-2016-demographic, hovy-yang-2021-importance, liandbamman2021}, we can also ask \emph{whose} language will be excluded. 

We begin with a summary of the handful of data sources used to construct training corpora for many language models and assumed to be of high quality (\S\ref{sec:pretraining_data}). The systematic authorship biases in these datasets motivate the study that follows, in which we replicate the quality filter from \citet{brown2020language}.  We apply this filter to a new dataset of U.S.~high school newspapers, augmented (via ZIP codes and counties) with demographic data from the U.S.~Census and the National Center for Education Statistics (\S\ref{sec:results}). We demonstrate that the filter has strong topical and stylistic preferences, and favors text from authors who originate from regions with better educational attainment, urban centers, larger schools, and higher valued homes.
% \dallas{I tripped on ``home values''. Maybe ``higher valued homes'' (or housing?)}

In sociolinguistics, the term \textbf{language ideology} refers to common (but often unspoken) presuppositions, beliefs, or reflections about language that justify its social use and structure \citep{doi:10.1146/annurev-linguistics-011718-011659}.  Our analysis begins to characterize the language ideology encoded in the quality filter used by \citet{brown2020language}, a representative of a wider set of filtering methods.  We also observe in \S\ref{sec:qual} that the filter is unaligned with other notions of quality familiar from human endeavors:  factuality ratings for news sources, standardized test scores, and literary awards.  Of course, these institutions hold their own language ideologies.  We argue that when constructing a corpus, one cannot avoid adopting some language ideology; the language ideology which is appropriate depends on the goals of the work, and one language ideology may conflict with another.  In short, there is no truly general-purpose corpus.

%  We also establish a methodology for investigating the language ideologies of future quality filters.
% Echoing others, we call for greater transparency and accountability around these choices, in both research and practice.

Our code and analysis is publicly available.\footnote{\url{https://github.com/kernelmachine/quality-filter}}

\section{Motivation:  Data Sources}
\label{sec:pretraining_data}
% \nascomment{retitled this section}
% \ekg{I took a stab at shortening this section, but I think others should take an additional pass and cut much more bc I think it is still too long bc this is not our main point... I'm even wondering if we should get rid of the sub sections? }
%Four primary datasets are used for model pretraining: \ekg{common crawl, wikiepdia, webtext and Books-- are these the right names?} Each reflects a presumption that vast swaths of text will overcome potential bias, yet all reflect certain assumptions and a lack of authorship diversity which mirror documented, problematic patterns in print archives detailed above. 
%\paragraph{\wikipedia}
%\ekg{to grab from slack Dallas' notes -- do we have a more systematic write up here?}
%\paragraph{\cc}

%\paragraph{\webtext}

%\paragraph{\books}

Across the many language models recently reported in the literature, the same small group of datasets have been routinely used as training corpora---Wikipedia, collections of books, and popular online articles (\S\ref{sec:lm_corpora}). These data are often treated as  exemplars of high quality text \citep{devlin2019bert, liu2019roberta, radford2019language, raffel2020exploring, brown2020language}. Although these datasets include text from many sources, extensive research suggests that the voices they represent are drawn from a relatively small, biased sample of the population, over-representing authors from hegemonic social positions.

\paragraph{\wikipedia}
\wikipedia~serves as a backbone for language models because of its scale, ease of use, permissive license, and goal of providing comprehensive coverage of human knowledge. However, although anyone can edit Wikipedia content, 
not everyone does. In practice, there are significant biases in Wikipedia authorship, content, and perspectives. 
For instance, despite efforts by Wikimedia, 
the site has been unable to resolve a persistent gender imbalance among its editors \citep{huang.2013,metawiki}. This imbalance is reflected in who gets written about, and how \citep{bamman2014unsupervised,graells2015first,wagner2015s}. There is also a pervasive urban bias; editors are less likely to come from rural areas, and coverage of these areas in Wikipedia tends to be more limited \citep{mandiberg.2020}.
Although coverage in English Wikipedia is not limited to those places where English is a majority language, an Anglo-American perspective dominates coverage.\footnote{For example, of the ten most frequently mentioned people in English \wikipedia, seven are U.S. Presidents, two are prominent figures in Christianity, and the only woman is the British monarch, Queen Victoria.}
Lastly, a relatively small number of people are responsible for most of the content \citep{panciera2009wikipedians,matei.2017}.  Wikipedia is thus less representative of language of the population than one might expect given its size and design.

\paragraph{Books}
Language models are also frequently trained on book corpora. BERT \citep{devlin2019bert} used
the Toronto BookCorpus \citep{7410368}, which consists of 7,185 self-published novels, a dataset criticized for copyright violation, poor quality control, imbalanced representation, and lack of documentation \citep{bandy.2021}.

\gptthree \citep{brown2020language} and The Pile \citep{gao2020pile} both use much larger corpora of books (although the former do not identify the source of this data). However, the Pile's books (also called \booksthree) are not a random selection. Rather, they appear to be drawn from a torrent file containing hundreds of thousands of copyrighted eBooks.

\booksthree is deserving of a more thorough investigation, but preliminary analyses reveal that the most prevalent authors in the corpus are American and British writers, especially of romance, mystery, and children's books (e.g., L.~Ron Hubbard, Danielle Steel, etc.). This pattern should be considered against the background of the American book publishing industry, which has been widely criticized as homogeneous \citep{leeandlow}.\footnote{This 2020 study found that Black people comprise only 5\% of the industry, and books by men tend to generate disproportionately more attention than those by women.} 

% \begin{figure}[t!]
%     \includegraphics[scale=0.5]{figures/owtc_distribution.pdf}
%     \caption{URL domains (top) of \reddit~submissions in \openwebtext, and the subreddits in which they appear (bottom). Each follow long-tailed distributions, with news-related content forming the overwhelming majority of content circulated on that platform.}
%     \label{fig:openwebtext}
% \end{figure}

\begin{table}[t!]
\centering
\small
% \resizebox{\textwidth}{!}{
\begin{tabular}{lcc}
%  & \multicolumn{2}{c}{\phighquality} \\

%  & & \multicolumn{4}{l}{Mean \phighquality} \\
%  & & \multicolumn{4}{l}{per school} \\
\toprule
 \bf URL Domain & \bf \# Docs  & \bf \% of Total Docs \\
\midrule 
bbc.co.uk        &     116K & 1.50\%\\
theguardian.com   &    115K & 1.50\%\\
washingtonpost.com  &   89K & 1.20\%\\
nytimes.com        &    88K & 1.10\%\\
reuters.com        &    79K & 1.10\%\\
huffingtonpost.com   &  72K & 0.96\%\\
cnn.com            &    70K & 0.93\%\\
cbc.ca             &    67K & 0.89\%\\
dailymail.co.uk    &    58K & 0.77\%\\
go.com            &     48K & 0.63\%\\

 \bottomrule
% \multirow{2}*{\bf M6} & \emph{Intercept} & $\phantom{-}$0.3191^{***}$ & \multirow{2}*{\emph{0.091}} \\
% & 2016 GOP Vote Share$^\dagger$^{\circ}  &  $-$0.0250$^{***}$ \\
% \midrule
% \multirow{3}*{\bf M7} & \emph{Intercept} & $\phantom{-}$0.3295$^{***}$& \multirow{3}*{\emph{0.382}} \\
% & Article Category &$\phantom{-}$0.0074$^{***}$\\
% & Num. Tokens &$\phantom{-}$0.0508$^{***}$ \\
\end{tabular}

% }
\caption{The most popular top-level URL domains in \openwebtext. Mainstream news forms the overwhelming majority of content in the dataset. Overall, just 1\% of the top-level URL domains in \openwebtext contribute 75\% of the total documents in the corpus.}
% \nascomment{should top row explain what's in parentheses (``Eval.'' like in the subheader for novel domains)?}}
\label{tab:owtc_distribution}
\end{table}

\paragraph{News and Other Popular Internet Content}

%Deviating from the Wikipedia plus books template
% OpenAI's open-source corpus of web content (\openwebtext) is extracted from content shared on Reddit. 
\citet{radford2019language}  %decided to use a different dataset, based on 
scrape text from the websites featured in popular Reddit submissions (i.e., those that received at least three upvotes) to construct the training data for \textsc{GPT-2}. As the original corpus is unavailable, we analyze its open-source replica, \openwebtext~\citep{owtc}.
% \input{tables/school_newspapers}

% Upvote systems shape identities and discourse on the Internet, by making certain content more visible to a wider array of users \citep{}. However, for many forums, the most popular content may be the most extreme or controversial \citep{} the more mainstrem In general, algorithms for incentivizing popular content on platforms like Reddit remain opaque \citep{}, but previous research has suggested that hateful \citep{} and controversial content, especially those that tend to target minority groups \citep{}, tend to be upvoted more. This is consistent with findings in other Internet platforms \citep{}.

% The content is also dominated by the political events occurring at the time of data collection\footnote{\openwebtext~was curated in 2019, and news related to Donald Trump tends to dominate the corpus.} and may disproportionately represent the formal writing styles used in national news outlets.

 We do not expect the corpus to represent a wide range of language variation.  Reddit users are mostly male, younger, and lean liberal, which influences the types of content shared on the platform.\footnote{As of 2016, 71\% of Reddit users are male, 59\% are between ages 18--29, and 43\% identify as liberal (vs.~19\% conservative): \url{https://pewrsr.ch/3FLbNL7}}  Viral media on the Internet assume similar characteristics; they tend to elicit awe, anger, or anxiety \citep{doi:10.1509/jmr.10.0353}, validate group identities \citep{doi:10.1177/1461444820958123}, and disseminate from users with authority  \citep{WEISMUELLER2022107150}.  

Indeed, we find that 1\% of the 311K unique top-level domains in \openwebtext contribute 75\% of documents in the corpus (Table \ref{tab:owtc_distribution}). 
%A few prominent news sources contribute a considerable proportion of articles in the corpus (Table \ref{tab:owtc_distribution}).
The most common websites in \openwebtext are internationally circulating British and American news outlets (e.g., \emph{BBC}, \emph{The New York Times}, \emph{The Washington Post}, \emph{The Guardian}), blogging platforms (e.g., \emph{Tumblr}, \emph{Blogspot}, or \emph{Medium}), sports content (e.g., \emph{ESPN}, \emph{SBNation}), and tech news (e.g., \emph{TechCrunch}, \emph{Wired}).  As expected, these links tend to appear on the most highly trafficked subreddits (e.g., \emph{/r/politics}, \emph{/r/worldnews}, \emph{/r/news}).

% The content is also dominated by the political events occurring at the time of data collection.\footnote{\openwebtext~was curated in 2019. In an 80M token sample of the corpus, we find that the name \emph{Donald Trump} is mentioned 9.9K times, and the term \emph{Brexit} is mentioned 3.0K times. \emph{Joe Biden}, on the other hand, is mentioned only 422 times \suchin{to update}}

% They also appear on subreddits prone to disseminating misinformation (\emph{/r/The\_Donald}, \emph{/r/conspiracy}; \citealt{isaac_conger_2021,10.1371/journal.pone.0225098}). 

These data are likely dominated by formal writing styles. Among news organizations, the adherence to slowly evolving style guides expresses specific linguistic standards \citep{froke_bratton_mcmillan_sarkar_schwartz_vadarevu_2020} and even geopolitical interests \citep{doi:10.1080/17512786.2012.674834}, which encourage rules about language use that can reinforce gender norms and racial hierarchies \citep{doi:10.1177/107769589404900209, doi:10.1177/0739532916634640}. 

% Slowly-evolving editorial standards do not reflect the fast-paced, dynamic nature of linguistic innovation \citep{Eisenstein2014DiffusionOL, Grieve2018MappingLI}.

%Echoing the issues raised with the above sources, various 
In general, a relatively homogeneous set of authors writes the majority of newswire \citep{grieco_2020}. Researchers find a striking lack of diversity in newsrooms and newspaper leadership.\footnote{As of 2018, racial minorities make up 37\% of the U.S. population, but only 17\% of staff and 13\% of leadership in U.S. newsrooms \citep{arana.2018}.} This may be compounded by the economic hardships aspiring journalists must incur,\footnote{In 2020, median salary for U.S. news analysis, reporters, and journalists was \$35,950, a slight decrease from 2012 after adjusting for inflation: \url{https://pewrsr.ch/3qCO75v}} which act as a filter for who can afford to be employed in newsrooms.

\paragraph{Summary}
Authors from specific, relatively powerful social positions produce a disproportionate amount of text in the core data sources of existing language models. These text sources favor privileged segments of the English-speaking population, including men, white populations,  communities of higher socio-economic status, and those harboring American and Western European historical, geopolitical, and cultural perspectives. By contrast, these corpora tend to be less inclusive of the voices of women and members of marginalized groups. Alternative perspectives, including those of people from rural areas, non-dominant gender, sexual, or racial identities, and counter-hegemonic vantage points, are less likely to be included, and thus less likely to influence models trained on this data.

Although formal, streamlined content like news or \wikipedia~articles may seem like desirable sources for high quality content, % (given that these source are from professional writers and editors covering prominent events), 
not all writing styles or substantive topics that might be relevant to language technologies and their user communities are represented in the resulting corpora. %Second, when the source content is disproportionately generated by certain (privileged) authors, these models 
%These sources then come to subjectively \textit{define}---rather than merely represent---what it means to be ``high quality.'' %This emerges in practice through the hiring and retention of writers and editors, the use of editorial standards which evolve slowly, and the adherence to style guides, which express both linguistic and geopolitical standards. 
When deployed, many of the technologies using language models trained on these data will face language that---despite being less formal, professional, or carefully edited---is no less high quality and is essential to the communicative lives of the people who use it. 

\section{Measuring the Language Ideology of the \openaifiltercaps }\label{sec:results}
% \section{Whose language do \emph{high quality} domains exclude?}
%\nascomment{this section should be retitled, probably making reference to the analysis using school newspaper data.  I don't know what ``positionality'' means precisely; readers won't, either.  I don't think we need special terminology for this.}

Empirically evaluating the full distribution of authors in the data sources from \S\ref{sec:pretraining_data} is difficult, due to their size, as well as their lack of metadata about each document's authors. We instead curate a new dataset of U.S. high school newspaper articles that varies both topically and along demographic variables that can be resolved using ZIP codes. 
Although we do not directly consider individual authors of these articles, this dataset is useful, in that it can be associated with extensive metadata at the level of individual newspapers.
We then analyze the behavior of a (replicated) quality filter on text from this dataset and discuss its implications. 
%patterns in the language (and authors) that a classifier which considers domains from \S\ref{sec:pretraining_data} as \emph{high quality} prefers when applied to the dataset. 
% In this section, we analyze if/how well quality filters pretrained on sources designated as \textit{high quality} represent a variety of  approaches to language, and we evaluate whether downstream exclusionary patterns emerge as a result.

\subsection{\schoolnewspapers}
\label{sec:data}

\paragraph{Background} Many U.S. schools produce a newspaper to give students journalism experience, to report on local news, to comment on national or global events, and to publish school-related material (e.g., announcements, campus life, student interviews, sports or honor rolls; \citealp{Gibson1961ASO}). The substantive content of school newspapers varies considerably, possibly due to their local audiences. %, so we expect considerable variation even among schools that used this template
Because a school's access to resources %(funding/teachers) 
is shaped by local income levels \citep{Betts2000EqualRE} and tied to student achievement \citep{doi:10.3102/00346543066003361}, we expect schools in wealthier areas (relative to poorer areas) to produce newspaper content that is more similar to the formal, professional texts that a quality filter is likely to classify as high quality. %We merge each school with its corresponding zipcode(s) and county-level income and demographic information from the U.S. Census. 

\paragraph{Collection} We collect articles from English-language U.S. school newspapers that used a common Wordpress template.\footnote{\url{SNOsites.com}} After retrieving 2483  schools who use this template, we scrape 1.95M articles from their respective newspaper sites (more details in \S\ref{sec:datasheet}). We retrieve article categories by extracting them from the article URL slugs. We then resolve each school to its ZIP code using the Google Maps Place API.\footnote{\url{https://developers.google.com/maps/documentation/places/web-service/search-find-place?hl=en}}
We restrict our dataset to articles from U.S. high schools. We only consider articles from 2010--2019, remove pages under the \emph{video}, \emph{photo}, or \emph{multimedia} categories, and remove schools that have less than 100 articles (which tend to contain scraping errors). The final corpus includes 910K articles, from 1410 schools, located in 1329 ZIP codes (552 counties) dispersed across all U.S. states (plus D.C.).   %Even among the school newspapers using this template, 

\paragraph{Limitations}
Our corpus is neither a random nor a representative sample of U.S. school newspapers. Instead, it represents schools that had sufficient Internet access, that elected to use a particular website template, and that maintain websites with retrievable archived content. The lack of representation in school newspaper leadership positions may influence which students contribute content to school newspapers  \citep{garcia_2021}. Educators also likely shape some articles, at least in part (though we expect them to be similarly affected by resource constraints). Finally, much of the content in these articles is specific to student concerns (e.g., sports, school events, campus culture, etc.), and the writing is, by definition, amateur. Nevertheless, because the corpus captures a wide range of content and geographical areas, it allows us to evaluate how  a quality filter handles real-world language variation, within a particular domain. %%It is dependent on schools having access to this online services, which may not be accessible in many communities; Long-tailed nature of school locations; 5) similar biases present in school newspapers as there are in national news \footnote{https://voices.aaja.org/index/2021/8/26/few-black-and-hispanic-students-are-editors-of-top-college-newspapers-survey-finds}.

%  However, it is precisely because these articles represent a diverse set of actual writing styles and topics that they are useful for our purpose. 

Using text from school newspapers introduces privacy concerns, especially since authors and subjects are minors. We therefore use this data only for evaluation purposes, and do \textbf{not} train (or release) any models on this data, or any raw text from the corpus. We do, however, release a Datasheet \citep{gebru2021datasheets} which documents the dataset's general characteristics and curation procedure (\S\ref{sec:datasheet}).

% \dallas{Actually, we can't release code that will work unless we extensively rewrite it, because of how the sites have changed.}

\subsection{The \openaifiltercaps}\label{sec:quality_filter_intro}

% \nascomment{subtle thing here.  calling it ``the quality filter'' makes it sound universal or like the only one that exists.  maybe better to call it ``OpenAI's quality filter" or something like that?}

% To measure the positionality  \nascomment{reword; is there an easier-to-understand name for the concept we're trying to operationalize/measure?} \dallas{agreed, this is not a standard usage of positionality} of datasets in \S\ref{sec:pretraining_data}, we implement a quality filter based on the approach proposed by \citet{brown2020language}
To investigate how quality correlates with various attributes of a newspaper, we re-implement the  \citealp{brown2020language} quality filter based on the description provided in the paper. The filter is a binary logistic regression classifier trained (using n-gram features) to distinguish between reference corpora (\booksthree, \wikipedia, and \openwebtext) and a random sample of \cc. 

% We replicate \citet{brown2020language}'s filter as closely as possible (see Appendix \S\ref{apx:quality_filter} for details). 

We replicate the filter as closely as possible using  \texttt{scikit-learn} \citep{scikit-learn}. To create the training data for the classifier, we sample 80M whitespace-separated tokens of \openwebtext, \wikipedia, and \booksthree each for the positive class, and 240M  whitespace-separated tokens of a  September 2019 \cc~snapshot for the negative class. We download the \cc~snapshot using code provided by \citet{Wenzek2020CCNetEH}. We perform a 100-trial random hyperparameter search, fixing only the hashing vectorizer and basic whitespace tokenization, following the implementation in \citet{brown2020language}.  See the search space and final hyperparameters of our replicated filter in \S\ref{sec:appendix_hps}. Our final classifier gets 90.4\% $F_1$ (91.7\% accuracy) on a set of 60M test tokens (30M held-out tokens from each class, or 72K documents from the negative class, and 33K from the positive class). We release code for training the quality filter and a demo of the trained filter.\footnote{\url{https://github.com/kernelmachine/quality-filter}}$^{,}$\footnote{\url{https://huggingface.co/spaces/ssgrn/gpt3-quality-filter}} We apply the quality filter to the \schoolnewspapers data, computing a quality score per document, which we denote \phighquality.

\subsection{Document-Level Analysis}\label{sec:topic}

\begin{figure}[t!]
    \centering
    \hspace*{-1cm}
\includegraphics[scale=0.6]{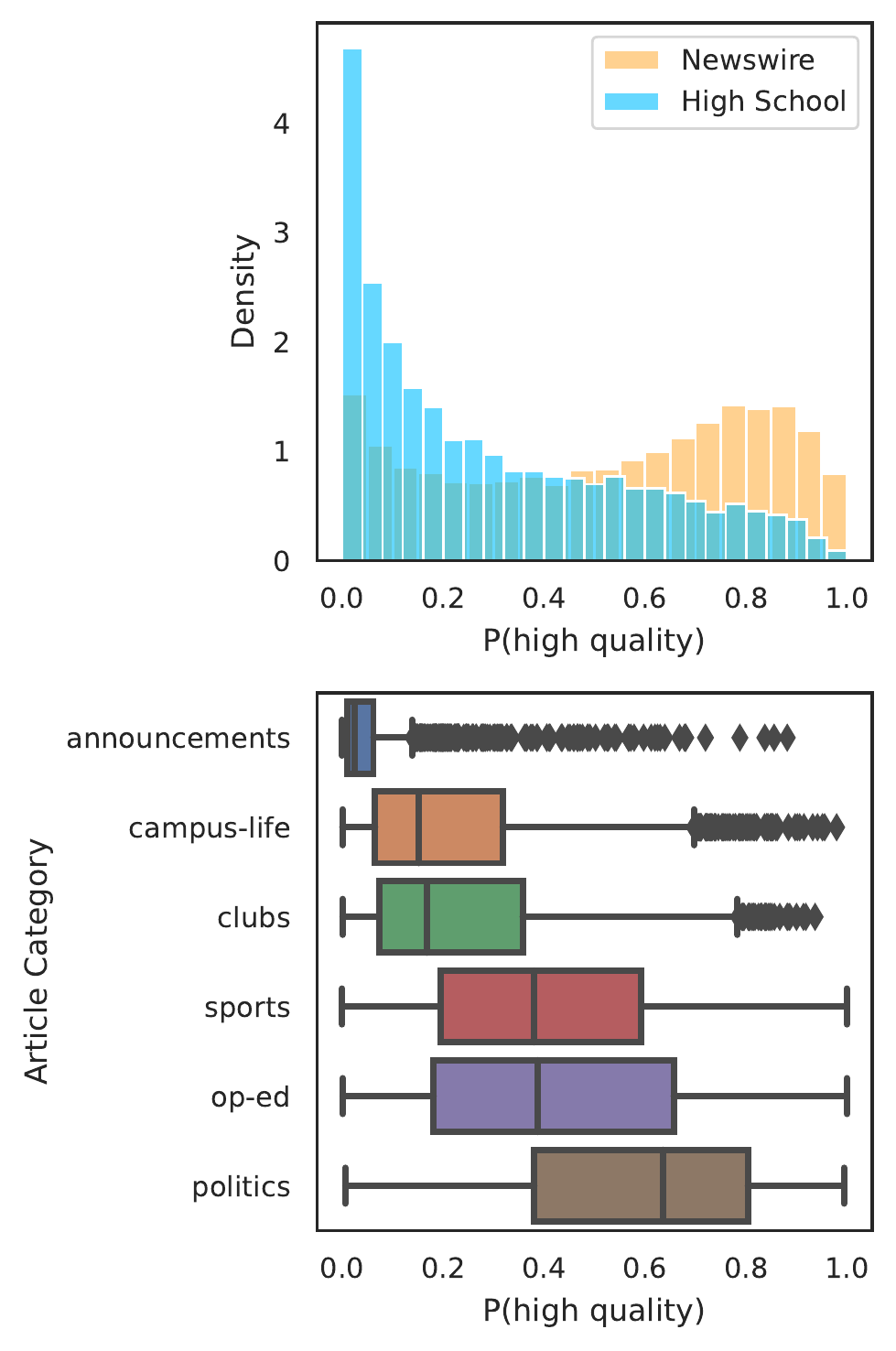}
    \caption{Scraped school articles tend to be considered lower quality by the \openaifilter than general newswire (histogram built from 10K random documents from each domain). This finding is consistent across a variety of categories, and more significant for certain ones (e.g., school announcements).}
    \label{fig:hsnews}
\end{figure}

We first explore document-level preferences of the filter. The \openaifilter~is more likely to classify high school newspaper articles as low quality,  compared to general newswire (Figure \ref{fig:hsnews}).\footnote{Here, the general newswire are articles from popular online news sources; see \S\ref{sec:qual} for data details.} This is unsurprising, since the training data for the \openaifilter included texts by professional journalists.  \S\ref{sec:example_articles} shows a random sample of  text from the dataset with high and low quality scores, illustrating differences in style and formality.  
% \nascomment{you need to explain a bit more about the topics here; basically the text in A.5 should be smoothed in here.  I had to guess what ``topic 0'' meant in the figure.}

More notably, controlling for article category (e.g., opinion pieces), we find that the \openaifilter has topical and stylistic preferences (discovered through exploratory data analysis). For topical features, we train an topic model (via Latent Dirichlet Allocation; \citealt{10.5555/944919.944937}) over opinion pieces with 10 topics using \texttt{scikit-learn}. We also consider whether documents contain first, second, or third person pronouns, and the length of the document. We then combine these features in a regression model to assess the effect of particular attributes on the quality score of a document, while controlling for others. 

The results of our regression are displayed in Table \ref{tab:regression_doc_level}. We find that certain topics have quite large effect sizes (see \S\ref{sec:topic_model} for the distribution of quality scores per topic). For example, documents entirely about Trump and the presidential election have quality scores 35 percentage points higher, on average, whereas documents about sports are 25 percentage points higher (relative to the omitted  topic about food). 
%In contrast, holiday- or school-related topics reduce the quality score by 6 percentage points and 4 percentage points, respectively.
Stylistically, the presence of first or second pronouns in a document decreases quality score by 5 percentage points, while a doubling of the number of tokens in a document increases the quality score by 9 percentage points.

\begin{table}[t!]
\centering
\small
% \resizebox{\textwidth}{!}{
\begin{tabular}{lc}
%  & \multicolumn{2}{c}{\phighquality} \\

\multicolumn{2}{c}{Dependent variable: \phighquality } \\
\multicolumn{2}{c}{Number of observations: 10K opinion articles}  \\

%  & & \multicolumn{4}{l}{Mean \phighquality} \\
%  & & \multicolumn{4}{l}{per school} \\
\toprule
 \bf Feature & \bf Coefficient \\
\midrule 
% \midrule[0.03em]
 \emph{Intercept} & $\phantom{-}0.471^{***}$\\
 Topic 5 (\emph{christmas, dress, holiday}) &   $-$0.056$^{***}$  \\
 Topic 2 (\emph{school, college, year}) &  $-$0.037$^{***}$ \\
 Topic 6 (\emph{student, school, class}) & $-$0.004$^{\phantom{***}}$ \\
 Topic 1 (\emph{people, just, like}) & $\phantom{-}$0.003$^{\phantom{***}}$ \\
  Topic 7 (\emph{movie, film, movies})  &  $\phantom{-}$0.062$^{***}$  \\
 Topic 3 (\emph{music, album, song}) &  $\phantom{-}$0.113$^{***}$ \\
  Topic 4 (\emph{people, women, media}) &  $\phantom{-}$0.197$^{***}$ \\
  Topic 9 (\emph{game, team, players}) &  $\phantom{-}$0.246$^{***}$ \\
 Topic 8 (\emph{Trump, president, election}) &  $\phantom{-}$0.346$^{***}$   \\
 Presence of first/second person pronoun &  $-$0.054$^{***}$ \\
 Presence of third person pronoun &    $\phantom{-}$0.024$^{\phantom{***}}$ \\
 log$_2$(Number of tokens) & $\phantom{-}$0.088$^{***}$ \\
 
 \midrule
 
 $R^2$ & 0.336\\
 adj. $R^2$ & 0.336 \\
% \multirow{2}*{\bf M6} & \emph{Intercept} & $\phantom{-}$0.3191^{***}$ & \multirow{2}*{\emph{0.091}} \\
% & 2016 GOP Vote Share$^\dagger$^{\circ}  &  $-$0.0250$^{***}$ \\
% \midrule
% \multirow{3}*{\bf M7} & \emph{Intercept} & $\phantom{-}$0.3295$^{***}$& \multirow{3}*{\emph{0.382}} \\
% & Article Category &$\phantom{-}$0.0074$^{***}$\\
% & Num. Tokens &$\phantom{-}$0.0508$^{***}$ \\
\bottomrule
\end{tabular}
% }
\caption{Regression of the quality score of an opinion piece in the \schoolnewspapers dataset, on document features.  We observe that political and sports-related topics, the lack of first and second person pronouns, and longer document lengths are associated with higher quality scores. We omit Topic 0 (\emph{food}, \emph{restaurant}, \emph{eat}) to avoid a saturated model. See \S\ref{sec:topic_model} for the distribution of quality scores per topic. $^*p <$ 0.05,  $^{**}p <$ 0.01, $^{***}p <$ 0.001.}
% \nascomment{should top row explain what's in parentheses (``Eval.'' like in the subheader for novel domains)?}}
\label{tab:regression_doc_level}
\end{table}

\subsection{Demographic Analysis}
\label{sec:demog}

Next, we examine whether the \openaifilter~prefers language from certain demographic groups over others.
We first check raw correlations between average quality scores (per newspaper) and features of interest. As in \S\ref{sec:topic}, we then combine the features in a regression model.

\paragraph{Demographic Features} As we note in \S\ref{sec:data}, we expect \emph{a priori} that content from schools located in wealthier, more educated, and urban areas of the U.S. will tend to have higher quality scores, relative to poorer, less educated, rural areas. Therefore, we consider demographic features that correspond to class, rural/urban divides, and school resources. 

For each school, we retrieve 2017--2018 school-level demographic data from the National Center for Education Statistics (NCES).\footnote{\url{https://nces.ed.gov/ccd/elsi/tablegenerator.aspx}} These include the number of students, student:teacher ratio, and indicators for charter, private, and magnet schools. We also retrieve the latest ZIP code- and county-level  demographic data from the 2020 U.S. Census.\footnote{\url{https://data.census.gov/cedsci/}} To measure the wealth of the corresponding ZIP code, we use median home values, and for educational attainment we use the percentage of college-educated adults. We also use Census data on the percent of rural population by county. Finally, we consider local political leanings, operationalized by GOP vote share in the 2016 Presidential election, using county-level data from the MIT election lab.\footnote{\url{https://electionlab.mit.edu/data}} 
We display full descriptions of features in our demographic analysis in \S\ref{sec:appendix_features}.

% DBC: Adding some minor details on variables
%\dallas{Adding some details on how we operationalize variables of interest}

% More specifically, from NCES, we obtain school characteristics (number of students, student:teacher ratio, and indicators for charter, private, and magnet schools), as well as student race demographics using the NCES categories. To measure the wealth of the corresponding ZIP code, we use median home values, and for educational attainment we use the percentage of college-educated adults, obtained from U.S. Census data, along with a categorization as urban or rural.\footnote{For all categorical variables, we drop one category from regressions to avoid a saturated model.} We also consider local political leanings, operationalized by the two-party vote share in the 2016 Presidential election. 

\begin{figure}[t]
    \centering
    \hspace*{-5mm}
    \includegraphics[scale=0.55]{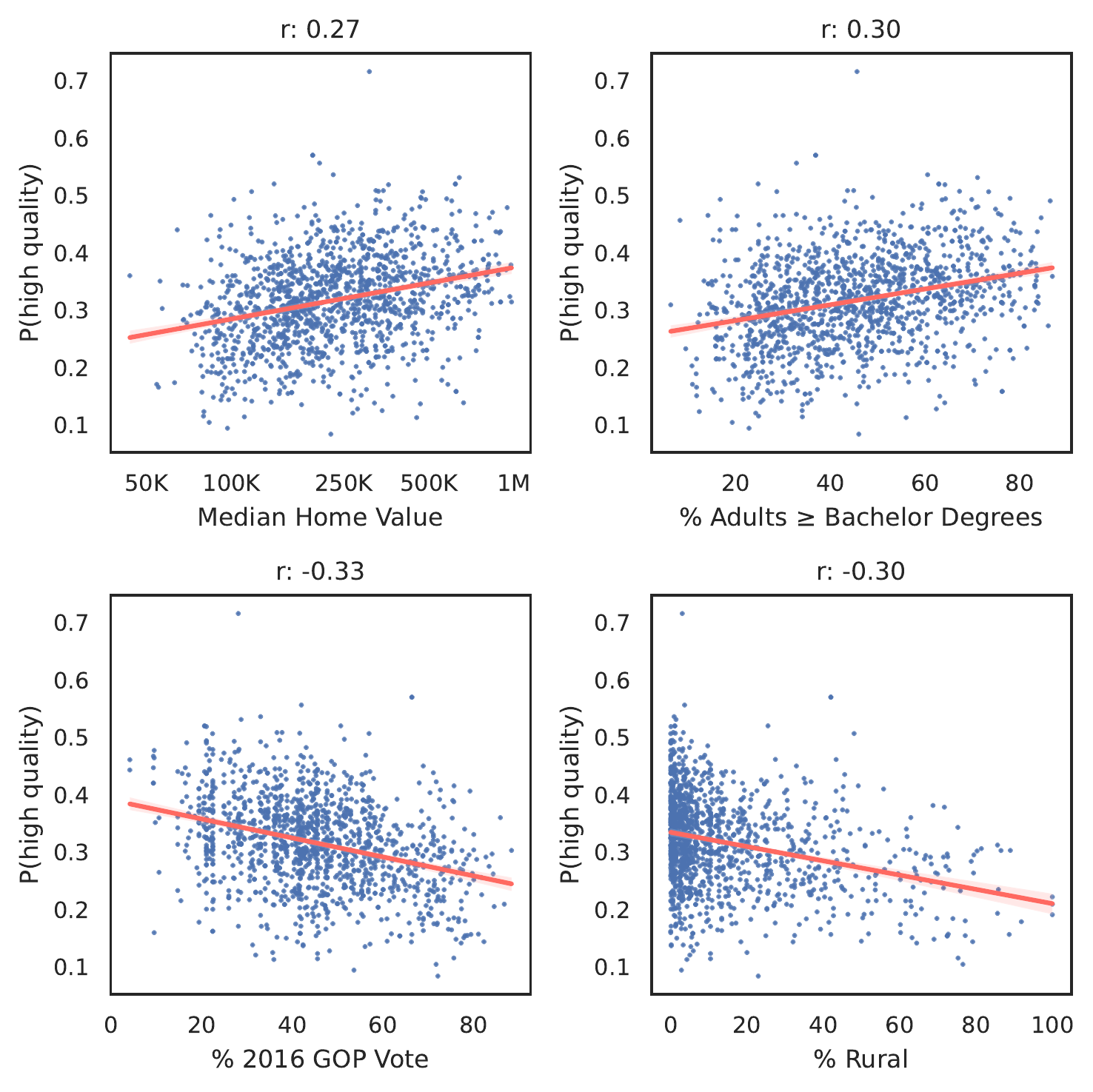}
    \caption{Scatter plots displaying correlations of select demographic features of a school's ZIP code or county with its average \phighquality.}
    \label{fig:basic_quality_stats}
\end{figure}

\paragraph{Correlation Analysis} To inform the variables we include in our regressions, we explore correlations between variables of interest and the average quality score of a school newspaper. Our analyses in  Figure \ref{fig:basic_quality_stats} suggest that our initial hypotheses hold: schools in wealthier, urban, and more educated ZIP codes, as well as those in Democrat-leaning counties, tend to have higher quality scores.

\paragraph{Data Preprocessing}\label{sec:dataprocess}  
Here, we use schools as the unit of analysis, and consider average quality score assigned to the school's articles as the dependent variable.  We only include those schools that could be matched to the NCES database, dropping schools which are missing school size, as well as those located in ZIP codes with \$1M or greater median home value, due to a census artifact.\footnote{The census data we use imposes an artificial upper bound on housing prices over \$1M.} Missing values for other features are imputed with the median value of that feature for the corresponding ZIP code, or (if necessary) county or state. For regressions, we log-transform school size, student:teacher ratio, and home values, using raw values for other features, to preserve interpretability. 
Our regression dataset includes 968 high schools, in 926 ZIP codes across 354 counties. All linear regressions are implemented with the \texttt{statsmodels} API.\footnote{\url{https://www.statsmodels.org/stable/index.html}} We release this anonymous dataset to support reproducibility.\footnote{ \url{https://github.com/kernelmachine/quality-filter}}

% As described in \S\ref{sec:data}, by resolving the zipcode in which each school resides, we are able to analyze demographic patterns of exclusion by the filter.

%We first
% \subsection{Results}

%\paragraph{Summary} Our descriptive analyses suggest that authors of school news articles located in wealthier, more educated, urban areas of the United States tend to produce higher quality content, according to \openaifilter. We confirm these analyses with a series of linear regressions in the next section.

%%%\subsubsection{Linear Regressions}\label{sec:regressions}\suchin{TODO (suchin): discuss data cleaning + update data stats after cleaning} \skd{Maybe put data cleaning in the appendix?}

%%\paragraph{Data} We limit our regression analyses to high school newspaper articles that fall within one of the 100 most-popular article categories. We aggregate all variables to the zipcode as our unit of analysis, which allows us to %%%We restrict our analysis to articles that are assigned one of the 100 most popular article categories among high schools. consider various zipcode- and county-level demographic features. We also track textual features %of the documents written by schools in each zipcode (e.g. average length of articles). Our final regression dataset includes 1.8K high schools in 1.7K zipcodes. %Tonormalize each feature, We log-transform features with large values (e.g. average home value) and produce z-scores for all features. We impute missing data with the median value of that feature for the corresponding county or (if necessary) state. 

\paragraph{Regression Analysis}
Because the variables identified above are correlated with each other, we use regression to estimate the effect of certain factors while controlling for others, with results shown in Table \ref{tab:main_regression}. Overall, home values, parental education, school size, public school status, and urban locations all show significant positive associations with quality scores. Thus, even controlling for financial resources, parental education, and other factors, articles from rural schools are still scored as significantly lower quality than those from urban schools.

Nevertheless, the effects, considered individually, are relatively modest. A 14 percentage point increase in percent urban population or a 17 percentage point increase in parental education (percent of adults with college degrees) correspond to a 1 percentage point increase in average quality score, as does a doubling of home values, or a quadrupling of school size. Average quality scores associated with public schools are  1.5 percentage points higher than private schools, controlling for other factors. Coefficients for charter schools, magnet schools, and student:teacher ratio are all sensible, though none are significant. Altogether, the combined effects of all these factors account for large differences in quality scores between wealthy, urban, educated locations, and poorer, rural, and less educated parts of the country.

\begin{table}[t!]
\centering
\small
% \resizebox{\textwidth}{!}{
\begin{tabular}{lc}
%  & \multicolumn{2}{c}{\phighquality} \\

\multicolumn{2}{c}{Dependent variable: \phighquality } \\
\multicolumn{2}{c}{Observations:  968 schools}  \\

%  & & \multicolumn{4}{l}{Mean \phighquality} \\
%  & & \multicolumn{4}{l}{per school} \\
\toprule
 \bf Feature & \bf Coefficient \\
\midrule 
% \midrule[0.03em]
 \emph{Intercept} & $\phantom{-}0.076^{\phantom{***}}$\\
 \% Rural & $-$0.069$^{***}$  \\
 \% Adults $\ge$ Bachelor Deg.  & $\phantom{-}$0.059$^{**\phantom{*}}$  \\
 log$_2$(Median Home Value) & $\phantom{-}$0.010$^{*\phantom{**}}$ \\ 
 log$_2$(Number of students) &   $\phantom{-}$0.006$^{*\phantom{**}}$ \\
 log$_2$(Student:Teacher ratio) &    $-$0.007$^{\phantom{***}}$ \\
 Is Public &  $\phantom{-}$0.015$^{*\phantom{**}}$   \\
 Is Magnet &  $\phantom{-}$0.013$^{\phantom{***}}$ \\
 Is Charter & $\phantom{-}$0.033$^{\phantom{***}}$ \\
 %\% Black Students &  $\phantom{-}$0.0021^{\phantom{***}} \\
 %\% Asian Students & $\phantom{-}$0.0048^{*\phantom{**}} \\
 %\% Native Students &   $\phantom{-}$0.0007^{\phantom{***}}  \\
 %\% Pacific Isl. Students &  $-$0.0018$^\phantom{***}$ \\
 %\% Mixed Students &  $\phantom{-}$0.0016$^{\phantom{***}}$  \\
 %\% Hispanic Students &  $-$0.0026$^{\phantom{***}}$ \\
 %Is  Urban & $\phantom{-}$0.0120^{*\phantom{**}}  \\
 \midrule
 $R^2$ & $\phantom{-}$0.140$^{\phantom{***}}$ \\
 adj. $R^2$ & $\phantom{-}$0.133$^{\phantom{***}}$  \\
% \multirow{2}*{\bf M6} & \emph{Intercept} & $\phantom{-}$0.3191^{***}$ & \multirow{2}*{\emph{0.091}} \\
% & 2016 GOP Vote Share$^\dagger$^{\circ}  &  $-$0.0250$^{***}$ \\
% \midrule
% \multirow{3}*{\bf M7} & \emph{Intercept} & $\phantom{-}$0.3295$^{***}$& \multirow{3}*{\emph{0.382}} \\
% & Article Category &$\phantom{-}$0.0074$^{***}$\\
% & Num. Tokens &$\phantom{-}$0.0508$^{***}$ \\
\bottomrule
\end{tabular}
% }
\caption{Regression of the average \phighquality~of a school in the \schoolnewspapers~dataset, on  demographic  variables. We observe that larger schools in educated, urban, and wealthy areas of the U.S tend to be scored higher by the \openaifilter. See \S\ref{sec:appendix_features} for more information on these features. $^*p <$ 0.05,  $^{**}p <$ 0.01, $^{***}p <$ 0.001.}
% \nascomment{should top row explain what's in parentheses (``Eval.'' like in the subheader for novel domains)?}}
\label{tab:main_regression}
\end{table}

\paragraph{Summary and Limitations} 

% Despite the original intentions of the \qualityfilter (e.g., to remove boilerplate HTML content), our analysis suggests that adopting datasets from \ref{sec:pretraining_data} as high quality references will leads to systematic biases agai

This analysis reveals an unintended consequence of the \openaifilter:
by attempting to exclude text that is less like mainstream news and Wikipedia, the filter reinforces a language ideology that text from authors of wealthy, urban, and educated  backgrounds are more valuable for inclusion in language model training data. These implicit preferences align with the attributes of authors that dominate the corpora from \S\ref{sec:pretraining_data}, which the filter considers to be high quality. 

Although most of the above findings are robust to alternate model specifications, the model ultimately only accounts for a relatively small amount of variance in quality scores. In addition, most of our features are taken from a single a point in time, and do not account for changing demographics over the period 2010--2019. Data errors could also arise due to how datasets were aligned (based on school name and ZIP code). These findings may not generalize to other domains (e.g., social media), and inclusion of additional features could affect these findings.
%We did considered the inclusion of racial demographics, using data from NCES, but found no significant effects, other than a positive association with the percentage of Asian students, which seemed to be driven at least in part by a correlation with home values.
For additional models which include vote share and racial demographics taken from NCES data, see \S\ref{sec:other_regressions}.

\section{Alignment with Other Notions of Quality}\label{sec:qual}

%The use of \wikipedia, newswire, books, and mainstream Internet content as baselines for high quality text is one of many possible ways to judge the quality of text. In fact, 
The \openaifilter purports to judge the quality of text. Humans, on the other hand, frequently judge the quality of text without the use of automated systems. In this section, we consider three forms of human evaluations: institutional awards to select books, fact-checkers' designated factuality of news outlets, and standardized test essays evaluated by human graders.  How well does the behavior of the \openaifilter map onto these other notions of quality?

\subsection{Data}

\paragraph{Factually (Un)reliable News}

To analyze the correspondence between the \openaifilter~and news factuality, we use the list provided by \citet{baly:2018:EMNLP2018} to identify a set of popular news sources from a broad range of factuality ratings and political leanings.\footnote{\citet{baly:2018:EMNLP2018} release a dataset of factual reliability and political leanings across news sources by scraping \url{NewsMediaBiasFactCheck.org}.} Using \texttt{Newspaper3k},\footnote{\url{https://newspaper.readthedocs.io/en/latest/}} we scrape and score 9.9K and 7.7K articles from high and low factuality news outlets, respectively. 

\paragraph{Essay Exams}

Next, to analyze the correspondence between the \openaifilter~and essay scores, we collect and score 12.1K participant essays from the \emph{Test Of English as a Foreign Language} (TOEFL) exam, a widely used  English language proficiency test \citep{Blanchard2013TOEFL11AC}. The TOEFL exam responses include official scores from exam readers, as well as each essay's prompt.

\paragraph{Award-Winning Literature} Finally, to analyze the correspondence between the \openaifilter~and literary awards, we select and score books from \booksthree and the Gutenberg corpus \citep{brooke-etal-2015-gutentag} that have won a Pulitzer Prize in various categories. We collected these data by scraping the publicly available list of recipients.\footnote{\url{https://www.pulitzer.org/prize-winners-categories}}

\begin{figure}[t!]
    \centering
    \hspace*{-0.5cm}
    \includegraphics[scale=0.55]{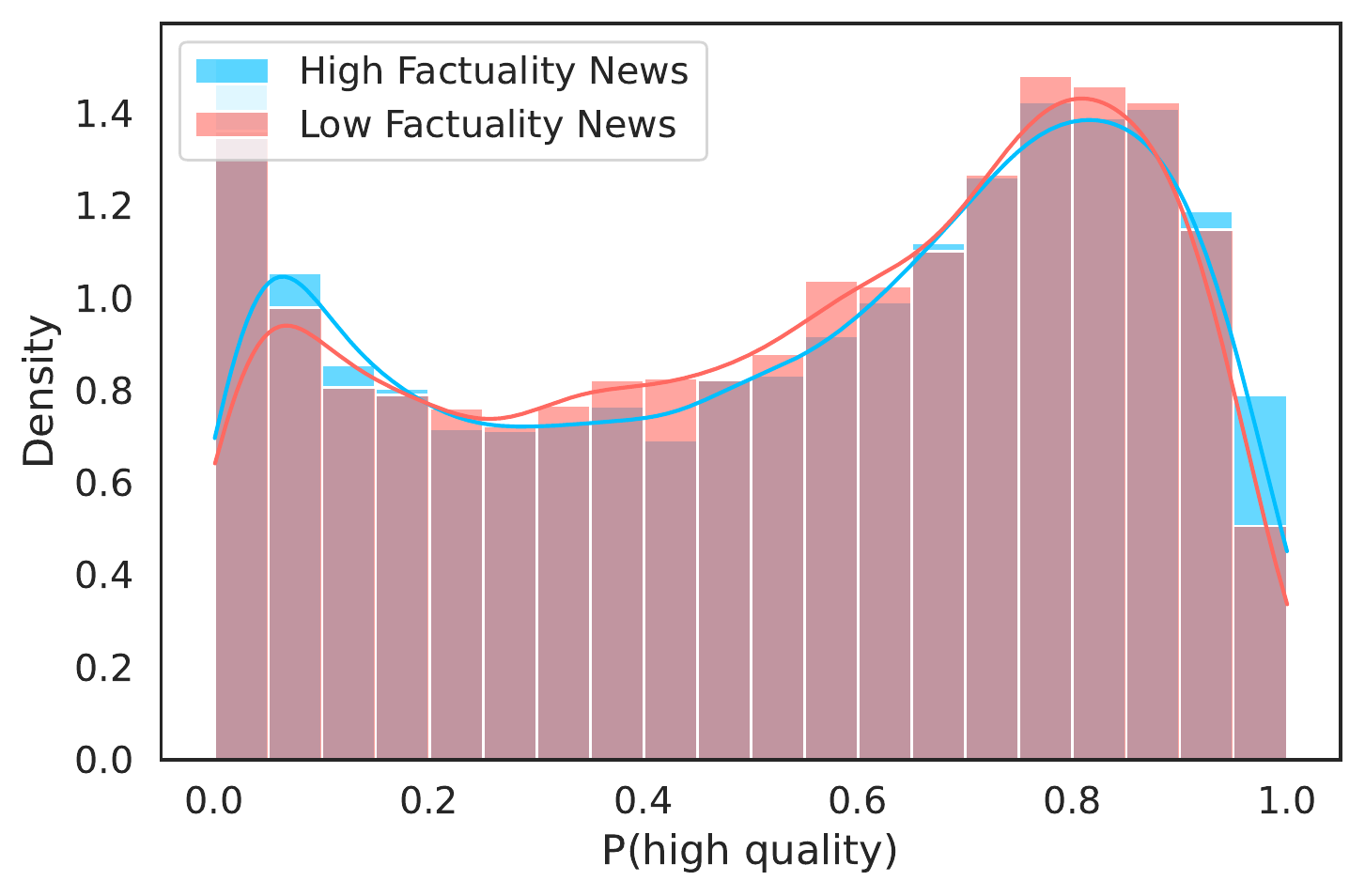}
    \caption{There is no difference in quality scores between articles written by news sources of high and low factual reliability.}
    \label{fig:fakenews}
\end{figure}

\subsection{Results}

If the filter aligns with news factuality, we would expect that articles from factually reliable sources would be rated as higher quality than those from factually unreliable ones. However, we find no difference in the quality distribution between articles from  high and low factuality news sources ($p=$ 0.085, two-way Kolmogorov-Smirnov test; Figure \ref{fig:fakenews}). Many factually unreliable news articles are considered high quality by the filter (\S\ref{sec:fake_news_examples}).

Turning to the TOEFL exam responses, we would expect that if the filter agrees with essay scores, higher scoring essays would receive higher quality scores. While essay scores are weakly correlated with quality scores (Pearson $r=$ 0.12, $p <$ 0.001), Table \ref{tab:toefl_regression} demonstrates that the essay's prompt is far more predictive of the essay's quality designation. For example, essays responding to a prompt (\emph{P4}) which asks participants to describe \emph{"...whether advertisements make products seem much better than they really are"} are much less likely to be filtered than all other prompts, including \emph{P6}, which asks participants to describe \emph{"...whether it is best to travel in a group"} (see \S\ref{sec:essay_prompts} for more details). The latter prompt tends to invoke personal experiences in the responses.

\begin{figure}[t!]
    \centering
    \hspace*{-5mm}
    \includegraphics[scale=0.50]{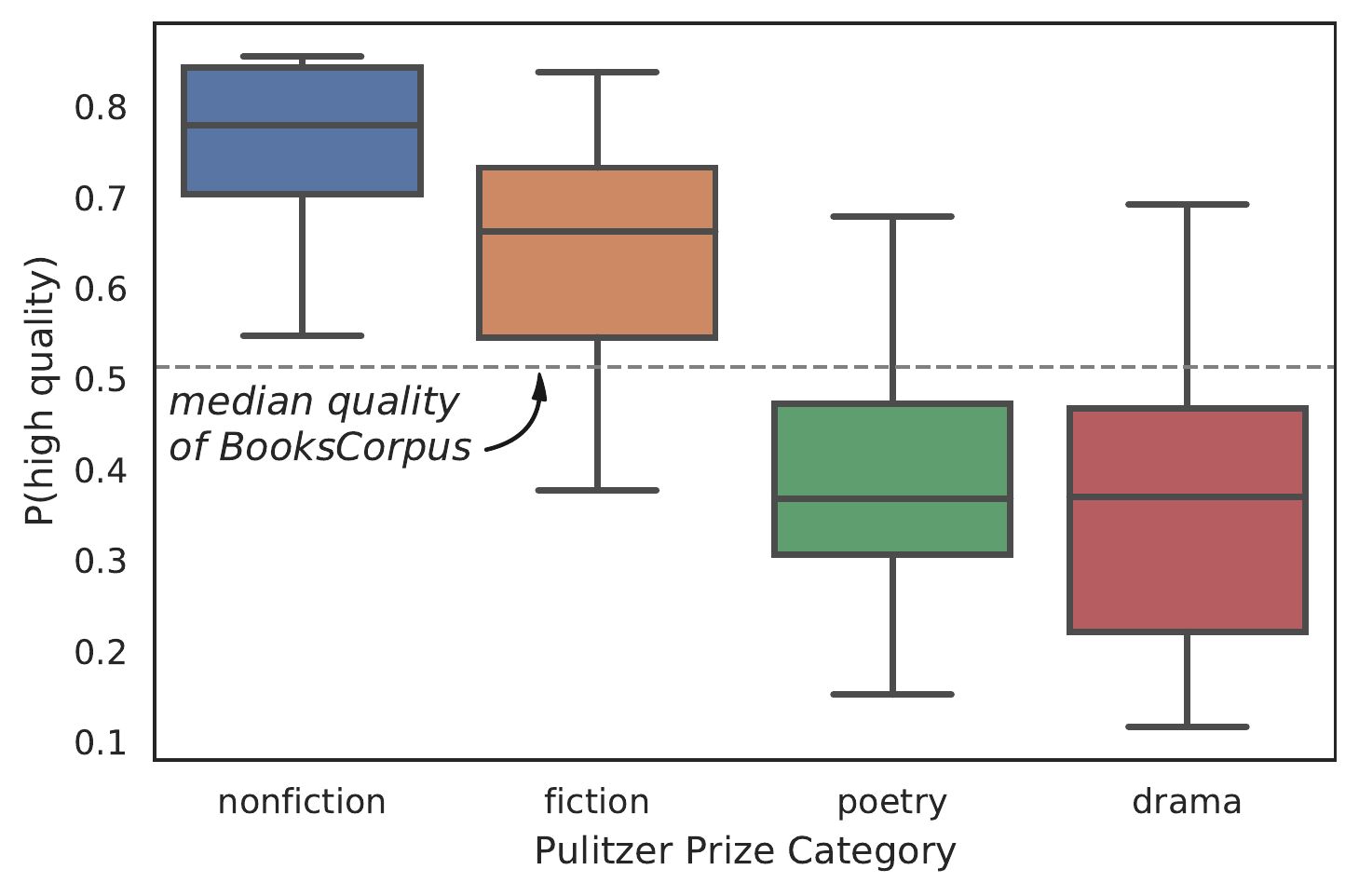}
    \caption{Among works that have won a Pulitzer Prize, the quality filter tends to favor nonfiction and longer fictional forms, disfavoring poetry and dramatic plays.}
    \label{fig:pulitzer_prize}
\end{figure}

Finally, if the filter aligns with literary awards, we would expect that most Pulitzer-Prize winning books would achieve high quality scores. On the contrary, quality scores vary heavily based on the genre (Figure \ref{fig:pulitzer_prize}). Poetry and drama are less favored by the filter relative to non-fiction, fiction, and even fan fiction (from the BookCorpus; \citealt{7410368}).  %\dallas{This section is excellently concise and clear}

\begin{table}[t!]
\centering
\small
% \resizebox{\textwidth}{!}{
\begin{tabular}{lc}
%  & \multicolumn{2}{c}{\phighquality} \\

\multicolumn{2}{c}{Dependent variable: \phighquality } \\
\multicolumn{2}{c}{Observations: 12.1K TOEFL exams}  \\

%  & & \multicolumn{4}{l}{Mean \phighquality} \\
%  & & \multicolumn{4}{l}{per school} \\
\toprule
 \bf Feature & \bf Coefficient \\
\midrule 
% \midrule[0.03em]
 \emph{Intercept} & $\phantom{-}$0.0631$^{***}$\\
 Low score &   $-$0.0414$^{\phantom{***}}$  \\
 High score &  $\phantom{-}$0.0339$^{\phantom{***}}$ \\
 Prompt 7 & $-$0.0283$^{***}$ \\
 Prompt 6 & $-$0.0204$^{***}$ \\
 Prompt 2 & $\phantom{-}$0.0068$^{***}$ \\
 Prompt 8 & $\phantom{-}$0.0346$^{***}$ \\
 Prompt 3 & $\phantom{-}$0.0880$^{***}$ \\
 Prompt 5 & $\phantom{-}$0.1470$^{***}$ \\
 Prompt 4 & $\phantom{-}$0.6745$^{***}$ \\
 \midrule
 
 $R^2$ & 0.712\\
 adj. $R^2$ & 0.711 \\
% \multirow{2}*{\bf M6} & \emph{Intercept} & $\phantom{-}$0.3191^{***}$ & \multirow{2}*{\emph{0.091}} \\
% & 2016 GOP Vote Share$^\dagger$^{\circ}  &  $-$0.0250$^{***}$ \\
% \midrule
% \multirow{3}*{\bf M7} & \emph{Intercept} & $\phantom{-}$0.3295$^{***}$& \multirow{3}*{\emph{0.382}} \\
% & Article Category &$\phantom{-}$0.0074$^{***}$\\
% & Num. Tokens &$\phantom{-}$0.0508$^{***}$ \\
\bottomrule
\end{tabular}
% }
\caption{Regression of the quality of a TOEFL exam essay on its assigned score and prompt.  While we observe some relationship between the score an essay receives and its quality score, the essay prompts themselves have significantly higher effect sizes. The highest quality essays come from Prompt 4, which asks participants to discuss products and advertisements. See \S\ref{sec:essay_prompts} for visualizations of distributions of quality across prompts and scores.  $^*p <$ 0.05,  $^{**}p <$ 0.01, $^{***}p <$ 0.001.}
% \nascomment{should top row explain what's in parentheses (``Eval.'' like in the subheader for novel domains)?}}
\label{tab:toefl_regression}
\end{table}

\paragraph{Summary}

Our analysis demonstrates that the \openaifilter conflicts with other standards of text quality. Of course, even the alternative standards we compare here are subject to their own language ideologies. Readers are more likely to trust news as factual if its political position aligns with their own \citep{mitchell_gottfried_barthel_sumida_2021}. English-language teaching pedagogies are rooted in ideologies about well-spokenness \citep{vanegas2016}. Literary awards favor white and male authors.\footnote{A 2016 study by the Columbia Journalism Review found that since 1918, 84\% of Pulitzer Prizes had been awarded to white authors, and 84\% to male authors: \url{https://www.cjr.org/analysis/100_years_of_data.php}.}  In general, any designation of text as high quality is subjective and influenced by sociopolitical context.

\section{Discussion} 
\label{sec:disc}

The above sections have demonstrated that automated filtering of text to build language modeling corpora may lead to counterintuitive or undesirable exclusion of sources. Because of the variety of use cases for language models and the broad range of text that could be appropriate for certain tasks, we suggest that there is no simple, universal standard for what should be considered high quality text. Indeed, there is a long history of privileging some people's  spoken language as better or more ``correct'' than others.  Researchers and practitioners of NLP who are aware of this history have the option to be intentional in their design of systems that, however implicitly, risk excluding the language of underprivileged identities or communities. 
% \dallas{nice rephrase!}

Some amount of selection in building corpora is inevitable. It is not possible to collect a uniform random sample of all written utterances.  However, our findings suggest that current selection methods are, for many purposes, flawed. Future work into alternative filtering criteria could be paired with investigations into the unintended consequences of their assumptions.

We do not believe that there is likely to be a single solution to this challenge. Indeed, the text that is best suited for training a model may depend on the application of that model. At a minimum, however, the NLP community could more carefully consider and clearly document the criteria by which text is being selected for inclusion. NLP practitioners could also be explicit about the reasons for using certain sources, even if those reasons are related to availability or empirical performance.  A collection of tests could also be deployed (and improved over time), to give a clear understanding of the implications of different choices of filters.  

More generally, we echo calls in the literature for more thoughtful and inclusive data collection \citep{Jo_2020,bender2021dangers,tanweer2021data}. This could include, but is not limited to a) intentionally curating data from people and viewpoints that are not otherwise well represented; b) including a greater diversity of genres; c) more nuanced or intentional exclusion criteria; d) more thorough interrogation of what text is being excluded; e) developing standard checks for prominent biases in inclusion; f) abandoning the notion of a general-purpose corpus.

\section{Ethical Considerations \& Limitations}

Our \schoolnewspapers dataset comes with many limitations, as  described in \S\ref{sec:data}. For example, the dataset contains sampling biases (e.g., it depends on use of a specific publication template), and the ZIP codes and counties are not uniformly spread across U.S. states. In general, our dataset likely captures neither the least resourced schools (which may not have access to online resources) in the United States, nor the wealthiest ones (who may have their own publication platforms). However, we speculate that an expanded corpus, which included writings from these schools, would demonstrate a continuation of trends we report in this paper.

While the text in our dataset varies considerably along topical, stylistic, and demographic variables, it is a niche domain; the text is a specific genre meant for local student consumption, its authors are U.S. students, and it thus primarily represents U.S.-centric cultural and political perspectives. We acknowledge that we also perpetuate some of the biases we identify, especially by working with English language text from the United States. We hope future work will extend this study of language ideologies to multilingual settings, other textual domains, and different sets of authors.

With respect to demographic variables, we merge census demographics with school-level data via ZIP codes or counties, which are imperfect identifiers of a school, since ZIP codes (and counties) may include multiple schools of varying resource levels. Moreover, tracking demographic variables and other author metadata, if deployed at scale, implies a certain level of invasive surveillance \citep{doi:10.1177/0003122417725865}. Future work may explore how to maintain the rights of authors as data subjects and producers while mapping demographic representation in large  corpora.

Finally, we did not seek consent from authors to scrape their articles. The ethical and legal norms around scraping public-facing web data, especially those produced by minors, are still in flux \citep{Fiesler2020NoRS}, and may not align with user perceptions of what constitutes fair use of online communications \citep{Williams2017TowardsAE}. For these reasons, we do not release the corpus of school newspaper articles, and only use it for analysis and evaluation. We only make available a dataset of demographic variables and quality scores per school, to support reproducibility.

% \suchin{One cannot take these results and make a "high quality" corpus}
% \suchin{Documenting lang. ideology, epistemology of science stuff}
% \suchin{Downstream users, no one-size-fits-all solution}
% \suchin{we are missing extremes, which may have a stronger effect? if we had to speculate, we would expect a continuation of trends, are missing the "bottom", it could be that we are underreporting. }
% \suchin{We are tying demographic vars by zipcode, which is imperfect}
% \suchin{Zipcodes may include multiple schools of varying resource level}
% \suchin{School Newspapers, artificial genre "for public consumption", not trying to engage with LM, more different than most things not more different than other things}
% \suchin{Dialects might have much stronger effect?}

% \suchin{Add GOP Vote Share Model + School-level model to appendix + Zip-code level model } 
% With respect to our recommendations in \S\ref{sec:disc}, the intentional inclusion of text from diverse sources may infringe on the rights of data subjects, despite good-faith efforts to protect their privacy \citep{Zimmer2010ButTD}. 

\section{Related Work}
\label{sec:related_work}

\paragraph{Language Ideologies} 
Language ideologies have been widely explored in the sociolinguistics literature \citep[][\emph{inter alia}]{10.2307/40971131, rosa_flores_2017, doi:10.1146/annurev-linguistics-011718-011659}. An ideology that promotes the inherent correctness, clarity, and objectivity of certain language varieties over others is a mechanism for linguistic discrimination \citep{doi:10.1146/annurev-linguistics-011718-011659, gal_2016, MacSwan2020-wz, Rickford2016-kf}. A salient example of such discrimination is the stigmatization of second-language speakers of English \citep{lindemann2005speaks}.

Language ideologies have an important, but often unacknowledged, influence on the development of NLP technologies \citep{blodgett-etal-2020-language}. For example, an ideology that distinguishes between \emph{standard} and \emph{non-standard} language variations surfaces in text normalization tasks \citep{van-der-goot-etal-2021-multilexnorm}, which tend to strip documents of pragmatic nuance \citep{baldwin-chai-2011-beyond} and social signals \citep{nguyen-etal-2021-learning}. Language on the Internet has been historically treated as a noisy variant of English, even though  lexical variation on the Internet is highly communicative of social signals \citep{eisenstein-2013-bad}, and varies considerably along demographic variables \citep{Eisenstein2014DiffusionOL} and community membership \citep{liandbamman2021}. Language ideologies also surface in tools for toxicity detection; for example, the classification behavior of the \textsc{Perspective API} (a popular hate speech detector) aligns with the attitudes of conservative, white, female annotators, who tend to perceive African-American dialects as more toxic \citep{sap2021annotators}. In this work, we examine the language ideology encoded in a widely used quality filter for text data selection.

\paragraph{Critiques of \emph{Laissez-Faire} Data Collection} We provide empirical evidence that \emph{laissez-faire} data collection (i.e., filtering large web data sources) leads to data homogeneity \citep{bender2021dangers}. As an alternative to \emph{laissez-faire} collection, \citet{Jo_2020} recommend drawing on institutional archival practices. However, we note that language ideologies are also prevalent (and may not be explicit) in institutional archives, which, for example, have preferred colonial voices over colonized ones when documenting historical events \citep{trouillot1995silencing, decker2013silence}.

% Previous studies have detailed other complex issues associated with collecting pretraining data, including consent \citep{Jo_2020}, copyright \citep{Kelli2020TheIO}, hatespeech \citep{Gehman2020RealToxicityPromptsEN}, and privacy \citep{Carlini2021ExtractingTD} concerns. Our findings and recommendations echo calls to move away from \emph{laissez-faire} data collection, and towards more careful curation.

\paragraph{Other Quality Filters}  Other definitions of text quality are used to create pretraining datasets, some of which do not rely on the datasets from \S\ref{sec:pretraining_data}. However, all techniques adopt language ideologies of what constitutes high quality text. \emph{Bad-word} filtering, which removes documents that contain certain stop-words,  disproportionately excludes language about and by minority groups \citep{dodge2021documenting}. Filtering Internet content for popularity \citep{radford2019language} leads to data homogeneity based on the characteristics of viral media and the composition of userbases in online forums (\S\ref{sec:pretraining_data}). Even lightweight filters \citep{aghajanyan2021htlm, rae2021scaling} put more emphasis on features like document length over factuality when determining what makes a document high quality. Any filtering method requires transparent justification and recognition of tradeoffs.

\paragraph{Downstream Behavior}  The behavior of language systems aligns with what we would expect from a language ideology that favors training data written by a narrow, powerful sector of society. For example, dialogue agents perform significantly worse when engaging in conversations about race \citep{10.1145/3173574.3173889} and with minority dialects of English \citep{10.3389/frai.2021.725911}. \gptthree~frequently resorts to use of stereotypes when minority groups are mentioned in its prompt \citep{abid.2021,blodgett2021sociolinguistically}.  \gptthree~is also prone to producing hate speech \citep{Gehman2020RealToxicityPromptsEN} and misinformation \citep{mcguffie2020radicalization}, which we would expect if its quality filter fails to distinguish the factual reliability of news sources in its training data (\S\ref{sec:qual}).  Concurrent to this work, \citet{gao2021empirical} show that aggressive data filtering with the \openaifilter degrades downstream task performance. A closer analysis of how the language ideologies in data selection lead to certain model behaviors is a rich area for future work. 
% This bias is complementary to other axes of bias in data t identifying hatespeech \citep{Gehman2020RealToxicityPromptsEN}, misinformation \citep{mcguffie2020radicalization}, and XXX in large pretraining datasets.  Previous work has suggested that \emph{laissez-faire} data collection leads to the prevalence of targeted hatespeech in pretraining data \citep{bender2021dangers}, our results show similarly problematic outcomes with respect to who implicitly participates in the production of these data. 

\section{Conclusion}

Using a new dataset of U.S. school newspapers, we find that the conventional, automated valuation of \wikipedia, newswire, \books, and popular Internet content as reference for high quality text implicitly favors content written by authors from larger schools in wealthier, educated, urban areas of the United States. Adopting this language ideology for text data selection leads to implicit, yet systematic and as-yet undocumented inequalities in terms of whose language is more likely to be included in training corpora. Although no single action will solve this complicated issue, data curators and researchers could be more intentional about curating text from underrepresented authors and groups, gathering sources from multiple genres and writing styles, and documenting their curation procedures and possible sources of exclusion.% \suchin{end with one sentence summary of section 6?}

\label{sec:additional_filters}

\clearpage

\section*{Acknowledgments}

This paper benefited from thoughtful feedback from a number of people: Emily M. Bender, Amber Boydstun, Timnit Gebru, Eun Seo Jo, Kelvin Luu, Lucy Li, Julian Michael, Amandalynne Paullada, Katharina Reinecke, Swabha Swayamdipta, Kelly Wright, and Kaitlyn Zhou. 

% We also benefited from early conversations and feedback from Su Lin Blodgett, Kelly Wright, and Katharina Reinecke.

\bibliographystyle{acl_natbib}
\bibliography{custom}
%\bibliography{anthology}

\clearpage
\appendix

\section{Appendix}
\label{sec:appendix}

\subsection{Language Model Training Corpora}
\label{sec:lm_corpora}

We display a list of popular language modeling corpora in Table \ref{tab:lm_overview}.

\begin{table*}
\centering
\small

\begin{tabular}{lp{10cm}r}
\toprule
 \bf Model & \bf Pretraining Data Sources  & \bf Citation  \\
\midrule 
 ELMo & 1B Word benchmark & \citep{peters-etal-2018-deep}  \\
   GPT-1 & BookCorpus  & \citep{radford2018language}  \\
  GPT-2 & WebText  & \citep{radford2019language}  \\
 BERT & BookCorpus + Wikipedia & \citep{devlin2019bert}   \\
  RoBERTa & BookCorpus + Wikipedia + CC-news + OpenWebText + Stories & \citep{liu2019roberta}  \\
   XL-Net & BookCorpus + Wikipedia + Giga5 + ClueWeb 2012-B + Common Crawl & \citep{NEURIPS2019_dc6a7e65}  \\
    ALBERT & BERT, RoBERTa, and XL-net's data sources & \citep{DBLP:conf/iclr/LanCGGSS20} \\ 

 T5 & Common Crawl (filtered)  & \citep{raffel2020exploring}  \\

 XLM-R & Common Crawl (filtered) & \citep{conneau2020unsupervised}\\

   BART & BookCorpus + Wikipedia  & \citep{lewis-etal-2020-bart}   \\

 GPT-3 & Wikipedia + Books + WebText (expanded) + Common Crawl (filtered) & \citep{brown2020language}    \\

 ELECTRA & BookCorpus + Wikipedia + Giga5 + ClueWeb 2012-B + Common Crawl & \citep{clark2020electra} \\
 
  Megatron-Turing NLG & The Pile + Common Crawl (filtered) + RealNews + Stories & \citep{kharya_alvi_2021}  \\
  
  Switch-C &  Common Crawl (filtered) & \citep{Fedus2021-jw}  \\
 Gopher & MassiveWeb + Books + Common Crawl (filtered) + News + GitHub + Wikipedia & \citep{rae2021scaling} \\
 \bottomrule

\end{tabular}

\caption{Overview of recent language models and their training corpora. All studies tend to draw from the same core data sources: Wikipedia, Books, News, or filtered web dumps.}

\label{tab:lm_overview}
\end{table*}

\subsection{Datasheet}
\label{sec:datasheet}

Our datasheet for the \schoolnewspapers dataset can be found here: \url{https://bit.ly/3rLrmwV}.

\subsection{Quality Filter Hyperparameters}\label{sec:appendix_hps}

\begin{table*}[t!]
    \centering
    \small
    \begin{tabular}{cc}
       \toprule
       \textbf{Computing Infrastructure} & 56 Intel Xeon CPU Cores\\ 
       \midrule
       \textbf{Number of search trials} & 100 \\
       \midrule
       \textbf{Search strategy} & uniform sampling \\
       \midrule
       \textbf{Best validation F1} & 90.4 \\
       \bottomrule
    \end{tabular}
    
    \vspace{3mm}\begin{tabular}{ccc}
        \toprule
        \textbf{Hyperparameter} & \textbf{Search space} & \textbf{Best assignment}\\
        \midrule
        regularization & \emph{choice}[L1, L2] & L1\\
        \midrule
        C & \emph{uniform-float}[0, 1] & 0.977778\\
        \midrule
        solver & 64 & liblinear\\
        \midrule
        tol & \emph{loguniform-float}[10e-5, 10e-3] & 0.000816\\
        \midrule
        ngram range & \emph{choice}["1 2", "1 3", "2 3"] & "1 2"\\ 
        \midrule
        random state & \emph{uniform-int}[0, 100000] & 44555 \\
        \midrule
        tokenization & whitespace & whitespace \\
        \midrule
        vectorization & hashing & hashing\\
        \midrule
        remove stopwords & \emph{choice}[Yes, No] & No \\
        \bottomrule
    \end{tabular}
    \caption{Hyperparameter search space and best assignments for our re-implementation of the \openaifilter.}
    \label{tab:classifier_hps}
\end{table*}

We display the hyperparameters of our logistic regression classifier (reproduction of the filter developed by \citealt{brown2020language}) in Table \ref{tab:classifier_hps}.

\subsection{Example Articles}
\label{sec:example_articles}

We display example articles and their quality scores in the \schoolnewspapers dataset in Table \ref{tab:newspaper_examples}.

\subsection{Topic Modeling}
\label{sec:topic_model}

\begin{figure*}[t!]
    \centering
    \includegraphics[scale=0.5]{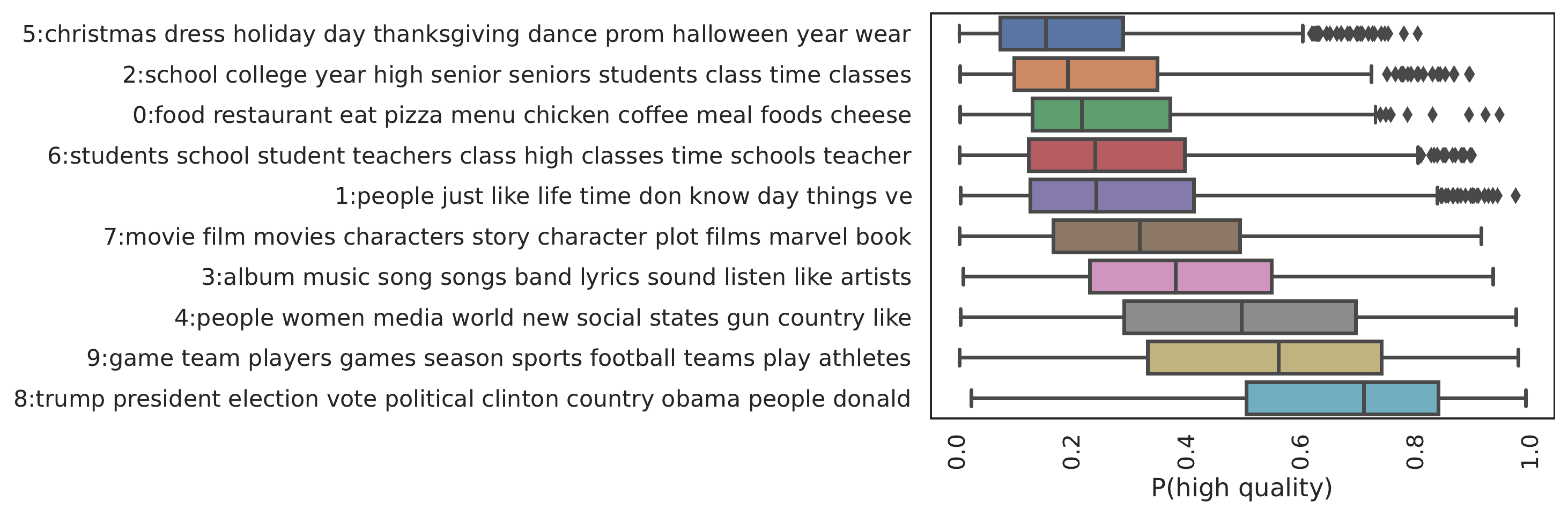}
    \caption{Considering 10K opinion pieces in \schoolnewspapers, we observe that the \openaifilter~prefers topics that are more prevalent in \wikipedia~or newswire. }
    \label{fig:topics}
\end{figure*}

See the quality distribution among topics for 10K opinion pieces in Figure \ref{fig:topics}.

\subsection{Demographic Features}
\label{sec:appendix_features}

We display a table of features we use in our demographic regression model in Table \ref{tab:demographic_features}.

\begin{table*}[t!]
\centering
\small
% \resizebox{\textwidth}{!}{
\begin{tabular}{lp{8cm}cc}
%  & \multicolumn{2}{c}{\phighquality} \\

%  & & \multicolumn{4}{l}{Mean \phighquality} \\
%  & & \multicolumn{4}{l}{per school} \\
\toprule
 \bf Feature & \bf Description  & \bf Level & \bf Source \\
\midrule 
 Is Charter & Is the school a charter school? & School & NCES database \\
 Is Private & Is the school a private school? & School &NCES database  \\
 Is Magnet & Is the school a magnet school?  & School & NCES database \\
  \% Black Students & \% students who identify as Black  & School & NCES database  \\
 \% Asian Students & \% students who identify as Asian  & School & NCES database \\
 %\% Native Students  & \% students who identify as Native American or Alaskan Native  & School & NCES database \\
 %\% Pacific Islander Students & \% students who identify as Pacific Islander  & School & NCES database   \\
 \% Mixed Students & \% students who identify as Mixed race  & School & NCES database     \\
 \% Hispanic Students & \% students who identify as Hispanic  & School & NCES database  \\
 Student:Teacher & Student-teacher ratio  & School & NCES database   \\
 School Size & Total number of students  & School & NCES database   \\
 Median Home Value & Median home value & ZIP code & Census  \\
 \% Adults $\ge$ Bachelor Deg.  & \% adults ($\ge$ 25 years old) with at least a bachelor's degree  & ZIP code & Census   \\
 \% Rural & Percent of a county population living in a rural area & County & Census  \\
 \% 2016 GOP Vote & Republican vote share in the 2016 presidential election & County & MIT Election Lab \\

 \bottomrule
% \multirow{2}*{\bf M6} & \emph{Intercept} & $\phantom{-}$0.3191^{***}$ & \multirow{2}*{\emph{0.091}} \\
% & 2016 GOP Vote Share$^\dagger$^{\circ}  &  $-$0.0250$^{***}$ \\
% \midrule
% \multirow{3}*{\bf M7} & \emph{Intercept} & $\phantom{-}$0.3295$^{***}$& \multirow{3}*{\emph{0.382}} \\
% & Article Category &$\phantom{-}$0.0074$^{***}$\\
% & Num. Tokens &$\phantom{-}$0.0508$^{***}$ \\
\end{tabular}
% }
\caption{Description of features we include in our demographic analyses.}
% \nascomment{should top row explain what's in parentheses (``Eval.'' like in the subheader for novel domains)?}}
\label{tab:demographic_features}
\end{table*}

\subsection{Additional Regressions}
\label{sec:other_regressions}

Here we include regressions results from two models with additional covariates.

We first consider race as a possible omitted variable, given the extent of school segregation in the U.S. \citep{reardon.2014}. NCES data provides the distribution of students by race for each school, using a particular set of racial categories, which comes with obvious limitations.
Nevertheless, we use the raw percentage scores provided as additional covariates in this model as a validity check.
We exclude the Native and Pacific Islander categories, due to imbalanced data and geographic concentration, as well as the white category, to avoid a saturated model. 

As shown in Table \ref{tab:regression_with_race}, the findings are nearly identical to the results in the main paper, with the exception that home values are no longer significant. The only racial category that shows a significant effect is Asian. However, we note a positive correlation between percentage of Asian students and median home values (Pearson $r=$0.32, $p <$ 0.001), suggesting that the variable for percentage of Asian students may be partially absorbing the effect of our measure of wealth.

Table  \ref{tab:gop_regression} shows the results for an alternate model which includes \% GOP vote share in the 2016 election. Once again, the results are very similar to the results in the main paper, although there is a strong (and significant) negative association between GOP vote share and quality scores, whereas the measures of home values and percent rural are no longer significant. 

The results for this model exemplify the difficulty of working with highly correlated variables. Given the strong association between GOP voters and rural areas, GOP vote share serves as an effective proxy for other variables of interest. However, because the results of the 2016 Presidential election were likely somewhat idiosyncratic, and because we find wealth and geography to be a more plausible explanation for differences in student writing than political preferences among their parents, we opt for the model without GOP vote share in the main paper.

% Please add the following required packages to your document preamble:
% \usepackage[normalem]{ulem}
% \useunder{\uline}{\ul}{}
\begin{table}[t!]
\centering
\small
% \resizebox{\textwidth}{!}{
\begin{tabular}{lc}
%  & \multicolumn{2}{c}{\phighquality} \\

\multicolumn{2}{c}{Dependent variable: \phighquality } \\
\multicolumn{2}{c}{Observations:  968 schools}  \\

%  & & \multicolumn{4}{l}{Mean \phighquality} \\
%  & & \multicolumn{4}{l}{per school} \\
\toprule
 \bf Feature & \bf Coefficient \\
\midrule 
% \midrule[0.03em]
 \emph{Intercept} & $\phantom{-}$0.134$^{\phantom{***}}$\\
 \% Rural & $-$0.073$^{***}$  \\
 \% Adults $\ge$ Bachelor Deg.  & $\phantom{-}$0.049$^{*\phantom{**}}$  \\
 log$_2$(Median Home Value) & $\phantom{-}$0.007$^{\phantom{***}}$ \\ 
 log$_2$(Number of students) &   $\phantom{-}$0.005$^{*\phantom{**}}$\\
 log$_2$(Student:Teacher ratio) &    $-$0.008$^{\phantom{***}}$ \\
 Is Public &  $\phantom{-}$0.020$^{*\phantom{**}}$   \\
 Is Magnet &  $\phantom{-}$0.013$^{\phantom{***}}$ \\
 Is Charter & $\phantom{-}$0.035$^{*\phantom{**}}$ \\
 \% Asian Students & $\phantom{-}$0.081$^{**\phantom{*}}$ \\
 \% Mixed Students &  $\phantom{-}$0.051$^{\phantom{***}}$  \\
 \% Black Students &  $-$0.009$^{\phantom{***}}$ \\
 \% Hispanic Students &  $-$0.020$^{\phantom{***}}$ \\
 %\% Native Students &   $\phantom{-}$0.0007^{\phantom{***}}  \\
 %\% Pacific Isl. Students &  $-$0.0018$^\phantom{***}$ \\
 %Is  Urban & $\phantom{-}$0.0120^{*\phantom{**}}  \\
 \midrule
 $R^2$ & $\phantom{-}$0.152$^{\phantom{***}}$ \\
 adj. $R^2$ & $\phantom{-}$0.142$^{\phantom{***}}$  \\
% \multirow{2}*{\bf M6} & \emph{Intercept} & $\phantom{-}$0.3191^{***}$ & \multirow{2}*{\emph{0.091}} \\
% & 2016 GOP Vote Share$^\dagger$^{\circ}  &  $-$0.0250$^{***}$ \\
% \midrule
% \multirow{3}*{\bf M7} & \emph{Intercept} & $\phantom{-}$0.3295$^{***}$& \multirow{3}*{\emph{0.382}} \\
% & Article Category &$\phantom{-}$0.0074$^{***}$\\
% & Num. Tokens &$\phantom{-}$0.0508$^{***}$ \\
\bottomrule
\end{tabular}
% }
\caption{Regression of the average \phighquality~of a school in the \schoolnewspapers~dataset, on  demographic  variables. As in the main paper, larger schools in educated and urban areas of the U.S tend to be scored higher by the \openaifilter. Asian is the only categorical race variable which shows a significant association (using data and categories taken directly from NCES). The association with home values is no longer significant, plausibly explained by a correlation between a higher proportion of Asian students and higher median home values. See \S\ref{sec:appendix_features} for more information on these features. $^*p <$ 0.05,  $^{**}p <$ 0.01, $^{***}p <$ 0.001.}
% \nascomment{should top row explain what's in parentheses (``Eval.'' like in the subheader for novel domains)?}}
\label{tab:regression_with_race}
\end{table}

\begin{table}[t!]
\centering
\small
% \resizebox{\textwidth}{!}{
\begin{tabular}{lc}
%  & \multicolumn{2}{c}{\phighquality} \\

\multicolumn{2}{c}{Dependent variable: \phighquality } \\
\multicolumn{2}{c}{Observations:  968 schools}  \\

%  & & \multicolumn{4}{l}{Mean \phighquality} \\
%  & & \multicolumn{4}{l}{per school} \\
\toprule
 \bf Feature & \bf Coefficient \\
\midrule 
% \midrule[0.03em]
 \emph{Intercept} & $\phantom{-}$0.248$^{**\phantom{*}}$\\
 \% Rural & $-$0.021$^{\phantom{***}}$  \\
 \% Adults $\ge$ Bachelor Deg.  & $\phantom{-}$0.067$^{**\phantom{*}}$  \\
 log$_2$(Median Home Value) & $\phantom{-}$0.003$^{\phantom{***}}$ \\ 
 log$_2$(Number of students) &   $\phantom{-}$0.006$^{**\phantom{*}}$ \\
 log$_2$(Student:Teacher ratio) &    $-$0.007$^{\phantom{***}}$ \\
 Is Public &  $\phantom{-}$0.017$^{*\phantom{**}}$   \\
 Is Magnet &  $\phantom{-}$0.009$^{\phantom{***}}$ \\
 Is Charter & $\phantom{-}$0.027$^{\phantom{***}}$ \\
 \% GOP vote share &  $-$0.114$^{***}$ \\
 %\% Asian Students & $\phantom{-}$0.0048^{*\phantom{**}} \\
 %\% Native Students &   $\phantom{-}$0.0007^{\phantom{***}}  \\
 %\% Pacific Isl. Students &  $-$0.0018$^\phantom{***}$ \\
 %\% Mixed Students &  $\phantom{-}$0.0016$^{\phantom{***}}$  \\
 %\% Hispanic Students &  $-$0.0026$^{\phantom{***}}$ \\
 %Is  Urban & $\phantom{-}$0.0120^{*\phantom{**}}  \\
 \midrule
 $R^2$ & $\phantom{-}$0.164$^{\phantom{***}}$ \\
 adj. $R^2$ & $\phantom{-}$0.157$^{\phantom{***}}$  \\
% \multirow{2}*{\bf M6} & \emph{Intercept} & $\phantom{-}$0.3191^{***}$ & \multirow{2}*{\emph{0.091}} \\
% & 2016 GOP Vote Share$^\dagger$^{\circ}  &  $-$0.0250$^{***}$ \\
% \midrule
% \multirow{3}*{\bf M7} & \emph{Intercept} & $\phantom{-}$0.3295$^{***}$& \multirow{3}*{\emph{0.382}} \\
% & Article Category &$\phantom{-}$0.0074$^{***}$\\
% & Num. Tokens &$\phantom{-}$0.0508$^{***}$ \\
\bottomrule
\end{tabular}
% }

\caption{Regression of the average \phighquality~of a school in the \schoolnewspapers~dataset, on  demographic  variables, including \% 2016 GOP Vote. We observe that including the political leaning of the county tends to wash out other variables, likely because partisan voting correlates heavily with other effects, like the urban/rural divide \citep{doi:10.1177/0002716217712696}. The only other covariates that stay significant are school size, parental education, and public (as opposed to private) schools. $^*p <$ 0.05,  $^{**}p <$ 0.01, $^{***}p <$ 0.001.}
% \nascomment{should top row explain what's in parentheses (``Eval.'' like in the subheader for novel domains)?}}
\label{tab:gop_regression}
\end{table}

\begin{table*}[t!]
\small
\begin{tabular}{llc}

\toprule
\bf ID & \bf Text                                                                                                                                                                                                                & \bf \phighquality \\ \midrule
P7        & It is more important for students to understand ideas and concepts than it is for them to learn facts.              & 0.04                         \\
P6        &  The best way to travel is in a group led by a tour guide.                                                            & 0.05                         \\
P1        & It is better to have broad knowledge of many academic subjects than to specialize in one specific subject.  & 0.07                         \\
P2        & Young people enjoy life more than older people do.                                                          & 0.08                         \\
P8        &  Successful people try new things and take risks rather than only doing what they already know how to do well.   & 0.10                         \\
P3        &  Young people nowadays do not give enough time to helping their communities.                                & 0.16                         \\
P5        &  In twenty years, there will be fewer cars in use than there are today.                                             & 0.22                         \\
P4        &  Most advertisements make products seem much better than they really are.                                   & 0.74                         \\ \bottomrule
\end{tabular}
\caption{TOEFL prompt IDs and their text, ordered by their quality score by \openaifilter.}
\label{tab:toefl_prompts}
\end{table*}

\begin{figure}[t!]
    \centering
    \includegraphics[scale=0.5]{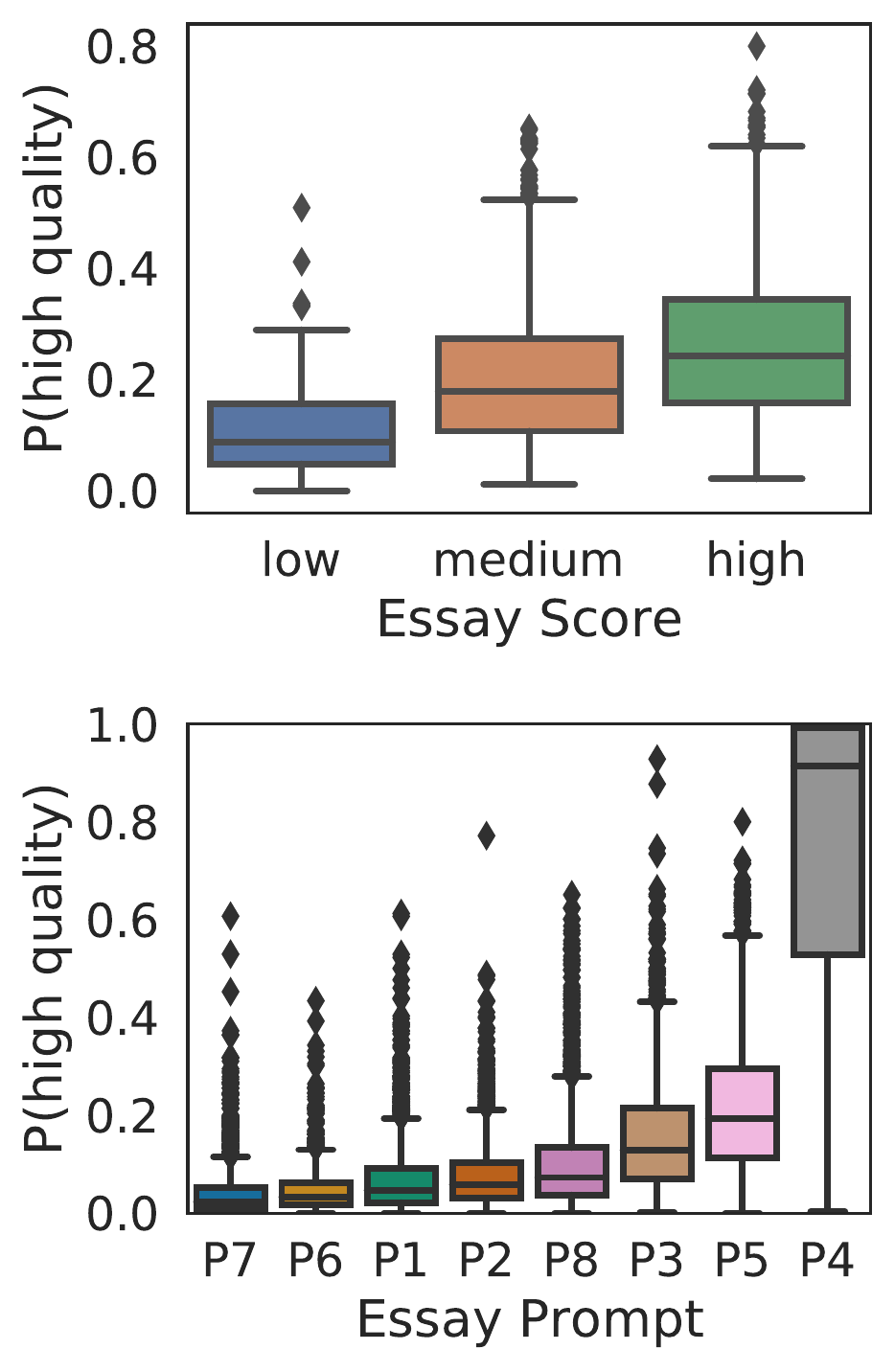}
    \caption{TOEFL exam score is weakly correlated with quality score across prompts (Pearson correlation; $r$=0.12 $\pm$ 0.05, $p \approx$ 0; top), but the essay prompt seems to be a much stronger indicator of quality scores than the exam scores are (bottom). }
    \label{fig:toefl_qual}
\end{figure}

\begin{table*}[t]
\centering
\small
\begin{tabular}{p{8cm}}
    % {>{\raggedright}p{11cm}p{6cm}}
    \toprule 
    \textbf{Category: Student-Life}\\
    \textbf{$P$(high quality) = 0.001} \\
    \midrule 
    
   \emph{As our seniors count down their final days until graduation, we will be featuring them each day. [REDACTED], what are your plans after graduation? To attend [REDACTED] in the fall and get my basics. Then attend the [REDACTED] program. What is your favorite high school memory? My crazy, obnoxious and silly 5th hour English with [REDACTED]. What  advice do you have for underclassmen? Pay attention, stay awake (I suggest lots of coffee), and turn in your dang work! You can do it, keep your head up because you are almost there!} \\
    \midrule
    \textbf{Category: News} \\
    \textbf{$P$(high quality) = 0.99} \\

    \midrule
    
    \emph{On Monday, September 3rd, Colin Kaepernick, the American football star who started the ``take a knee'' national anthem protest against police brutality and racial inequality, was named the new face of Nike’s ``Just Do It'' 30th-anniversary campaign. Shortly after, social media exploded with both positive and negative feedback from people all over the United States. As football season ramps back up, this advertisement and the message behind it keeps the NFL Anthem kneeling protest in the spotlight.} \\
    \bottomrule
\end{tabular}

\caption{
    Examples of high school news paper articles from \schoolnewspapers. Many of the articles in student-life category, and similar, rated lower quality have very different styles from documents rated high quality.
} 
\label{tab:newspaper_examples}
\end{table*}

\subsection{Low Factuality News Considered High Quality}
\label{sec:fake_news_examples}

We display example low factuality news articles that are assigned high quality scores by the \openaifilter in Table \ref{tab:fake_news_examples}.

\begin{table*}[t]
\centering
\small
\begin{tabular}{p{8cm}}
    % {>{\raggedright}p{11cm}p{6cm}}
    \toprule 
    \textbf{Article from \emph{http://en-volve.com}}\\
    \textbf{$P$(high quality) = 0.93} \\
    \midrule 
\emph{The German government has effectively began the process of eliminating the unvaccinated by starving them to death by pushing grocery stories to ban unvaccinated residents from buying essential food items...The pressure on the unvaccinated grows and grows!...}  \\  
    \midrule
    \textbf{Article from \emph{http://www.censored.news}} \\
    \textbf{$P$(high quality) = 0.98} \\

    \midrule
    \emph{The provisional number of births in the U.S. was 3,605,201 in 2020. That is the lowest number of births in the United States since 1979, according to the Centers for Disease Control. 2020 also had the lowest fertility rate since the government started tracking births in 1902. And don’t blame the so-called “pandemic.”...we’re learning in 2021 that intelligent people succumb to government psy-ops. But critical thinkers understood immediately that something was very wrong with all the COVID-19 stuff. Plus many among the global elite continually and openly gloat about their desire to cull the masses. Bill Gates isn’t even coy about his desires...} \\
    \bottomrule
\end{tabular}

\caption{
    Examples of news from low factuality sources (as identified by \url{MediaBiasFactCheck.com}) rated high quality by \openaifilter, but contain COVID disinformation.
} 
\label{tab:fake_news_examples}
\end{table*}

\subsection{TOEFL Exam Responses}
\label{sec:essay_prompts}

We display the distribution of quality scores against prompts and essay scores in the TOEFL exam dataset in Figure \ref{fig:toefl_qual}. We display the prompts of this dataset in Table \ref{tab:toefl_prompts}.

\end{document}